\documentclass[11pt, a4paper]{article}
\usepackage[utf8]{inputenc}
\usepackage[T1]{fontenc}
\usepackage{lineno}
\usepackage{graphicx}
\usepackage{geometry}
\usepackage{array}
\usepackage{booktabs}
\usepackage{textcomp}
\usepackage{algorithm}
\usepackage{algpseudocode}
\usepackage{phaistos}
\usepackage{graphicx}
\usepackage{subcaption}
\usepackage{caption}
\usepackage{xtab}
\usepackage{multicol}
\usepackage{tabularx}
\usepackage{longtable}
\usepackage{adjustbox}
\usepackage{multirow}
\usepackage{svg}
\usepackage{amssymb}
\usepackage{amsmath}
\usepackage{xspace}
\usepackage{authblk}
\usepackage{dsfont}
\usepackage{lineno}
\usepackage{makecell}
\usepackage{siunitx}
\usepackage{makecell}
\usepackage[labelfont={bf,it},textfont=it,singlelinecheck=false]{caption}
\usepackage{siunitx}
\DeclareSIUnit\pixel{px}
\usepackage[numbers,compress]{natbib}

\usepackage{hyperref}
\hypersetup{
    colorlinks=true,
    linkcolor=blue,
    filecolor=magenta,      
    urlcolor=cyan,
    pdftitle={Overleaf Example},
    pdfpagemode=FullScreen,
    }
\date{}
\DeclareUnicodeCharacter{2032}{\ensuremath{'}}
\DeclareUnicodeCharacter{00B0}{\ensuremath{^\circ}} 
\DeclareUnicodeCharacter{2212}{-}

\DeclareMathOperator{\Dead}{{\textit{Dead coral}}}

\DeclareMathOperator{\AcroporeB}{\textit{Acropora Branching}}
\DeclareMathOperator{\AcroporeD}{\textit{Acropora Digitate}}

\DeclareMathOperator{\AcroporeT}{\textit{Acropora Tabular}}

\DeclareMathOperator{\NoAcroporeE}{\textit{Non-acropora Encrusting}}

\DeclareMathOperator{\NoAcroporeM}{\textit{Non-acropora Massive}}

\DeclareMathOperator{\Syringodium}{{\textit{Syringodium isoetifolium}}}
\DeclareMathOperator{\Thalassodendron}{{\textit{Thalassodendron ciliatum}}}

\DeclareMathOperator{\AlgaeA}{\textit{Algal Assemblage}}
\DeclareMathOperator{\Algae}{\textit{Algae}}

\DeclareMathOperator{\SeaC}{{\textit{Sea cucumber}}}

\DeclareMathOperator{\Sand}{\textit{Sand}}
\DeclareMathOperator{\Rubble}{\textit{Rubble}}

\begin{document}
\let\WriteBookmarks\relax
\renewcommand{\floatpagefraction}{0.8}
\renewcommand{\textfraction}{0.1}
\renewcommand{\topfraction}{0.9}
\renewcommand{\bottomfraction}{0.8}

\title{A drone-based framework for coral habitat mapping via weakly supervised segmentation} 
\author[1,2, *]{Matteo Contini}
\author[1]{Victor Illien}
\author[4]{Sylvain Poulain}
\author[3]{Serge Bernard}
\author[4]{Julien Barde}
\author[1]{Sylvain Bonhommeau}
\author[2]{Alexis Joly}

\affil[1]{IFREMER Délégation Océan Indien (DOI), Le Port, 97420, La Réunion, France, Rue Jean Bertho}
\affil[2]{INRIA, LIRMM, Université de Montpellier, CNRS, Montpellier, 34000, France}
\affil[3]{CNRS, LIRMM, Université de Montpellier, Montpellier, 34000, France}
\affil[4]{UMR Marbec, IRD, Université de Montpellier, CNRS, Ifremer, Montpellier, 34000, France}

\affil[*]{corresponding author: Matteo Contini (matteo.contini@ifremer.fr)}

\maketitle

\begin{abstract}

Obtaining pixel-level annotations over large spatial extents remains a major bottleneck for deploying machine learning in ecological applications.
Here we present a multi-scale weakly supervised semantic segmentation (WSSS) framework that enables training high-resolution segmentation models from dense, classification-based outputs.
Our method combines fine-scale, multi-label predictions from underwater imagery with broad-coverage aerial data.
We convert these point-level classifications into coarse supervision masks that can be used to train a semantic segmentation model on Unmanned Aerial Vehicle (UAV) orthophotos.
A second training step using the model’s own refined predictions is then used to further improve spatial accuracy without requiring additional annotations.
We demonstrate the approach on coral reef imagery, enabling large-area segmentation of coral morphotypes and illustrating its flexibility in integrating new classes.
The final model achieves 86.07\% pixel accuracy and 52.23\% mean Intersection over Union (mIoU) on manually annotated reef zones, demonstrating that accurate large-scale coral segmentation can be obtained without pixel-level annotations.
By bridging image classification and segmentation across scales and modalities, this method provides an efficient solution for deploying segmentation models in settings where annotations are unavailable and opens opportunities for scalable, efficient monitoring in ecology and beyond.

\vspace{0.5cm}

\end{abstract}

\section{Introduction} 
\label{sec_introduction}

Coral reefs are vital marine ecosystems that support biodiversity, protect coastlines and sustain fisheries and tourism industries \cite{Hoegh-Guldberg2, rogers, bruckner2002life, hidayati2022importance}. 
However, they are increasingly threatened by climate change, ocean acidification, pollution and human activities, leading to widespread degradation and habitat loss \cite{hughes2017global, hughes2003climate, van2011chemical}. 
Effective reef monitoring at large spatial and temporal scales is crucial for conservation and management efforts. 
Although traditional diver-based surveys provide high-resolution and quality data, they are limited in coverage, labour intensive and costly \cite{cardenas2024systematic}. 
Recent advances in remote sensing technologies, particularly unmanned aerial vehicles (UAVs) and autonomous surface vehicles (ASVs), offer scalable solutions for large-scale coral reef monitoring \cite{MISIUK2024108599, contini2025seatizen}.
UAVs enable rapid aerial assessments of extensive reef areas \cite{borja2024innovative}, while ASVs provide fine-scale underwater observations \cite{gogendeau2025autonomous}.
Additionally, spatial interpolation of ecological indicators from plot-based surveys has been proposed as a cost-effective strategy for rapid biodiversity assessment \cite{broudic2025mapping}.
However, integrating data across these different scales remains a key challenge.

Deep learning-based image analysis has transformed marine monitoring applications, allowing automated classification of complex environments \cite{talpaert2023geoai, tuia2022perspectives, besson2022towards}. 
Segmentation-based image analysis has transformed marine monitoring applications, allowing automated classification of underwater complex coral environments \citep{alonso2019coralseg, raine2022point, zhong2023combining, zheng2024coralscop}. 

Fully supervised semantic segmentation models require pixel-wise annotations, but producing such detailed labels for large-scale reef monitoring is labour-intensive and time-consuming \citep{sauder2024scalable, sauder2025coralscapes}. 
Moreover, while broad habitat classes such as sand, coral, or seagrass can be reliably identified on UAV images \citep{doukari2022overcoming, qu2024gan, kvile2024drone}, distinguishing between different coral morphotypes is challenging due to the limited resolution of UAV imagery that often does not provide enough discriminative detail to differentiate fine-scale morphological variations.

The problem is further exacerbated by the visual complexity of benthic scenes and the presence of optical distortions such as caustics (patterns of light and shadow caused by the refraction of sunlight through the wavy water surface) that produce noisy and ambiguous images where object boundaries are difficult to discern.
This challenge demands alternative approaches that reduce annotation efforts while preserving annotation accuracy.

Weakly supervised semantic segmentation (WSSS) applied to remote sensing images offers a promising solution by leveraging weak labels, such as image-level annotations, bounding boxes, or rough segmentation masks, instead of manually labelled pixel-wise ground truth masks \cite{10330561, wang2023weakly}.
This technique reduces the need for extensive manual annotation by training models on approximate labels (human or machine-generated) and learning to predict masks from these weak annotations.
Recent WSSS research has improved pseudo-label generation through the use of foundation models and refinement strategies, including SAM-assisted pseudo-mask generation \cite{kweon2024sam} and CLIP-based single-stage segmentation \cite{zhang2024frozen}. 
Related efforts have also emerged in remote sensing, where weak supervision has been combined with SAM-based refinement to improve land-cover segmentation \cite{chen2025weakly}. 
In marine ecology, weak supervision has been explored for underwater segmentation using sparse point labels \cite{raine2026ai} and image-level labels for coarse seagrass mapping \cite{raine2024image}. 
However, all these approaches operate within a single modality and spatial scale, assuming that weak annotations are derived from the same image domain as the target segmentation task.
As a result, they do not explicitly address scenarios involving cross-modal supervision or differences in spatial resolution between the source of weak labels and the target imagery.
In contrast, our setting requires leveraging fine-scale underwater observations to supervise segmentation on large-scale aerial imagery, introducing two key challenges that are not addressed in standard WSSS frameworks: (i) transferring knowledge across modalities with distinct visual characteristics and (ii) aligning sparse, pointwise predictions with high-resolution orthophotos.

\begin{figure}[ht]
    \centering
    \includegraphics[width=\textwidth]{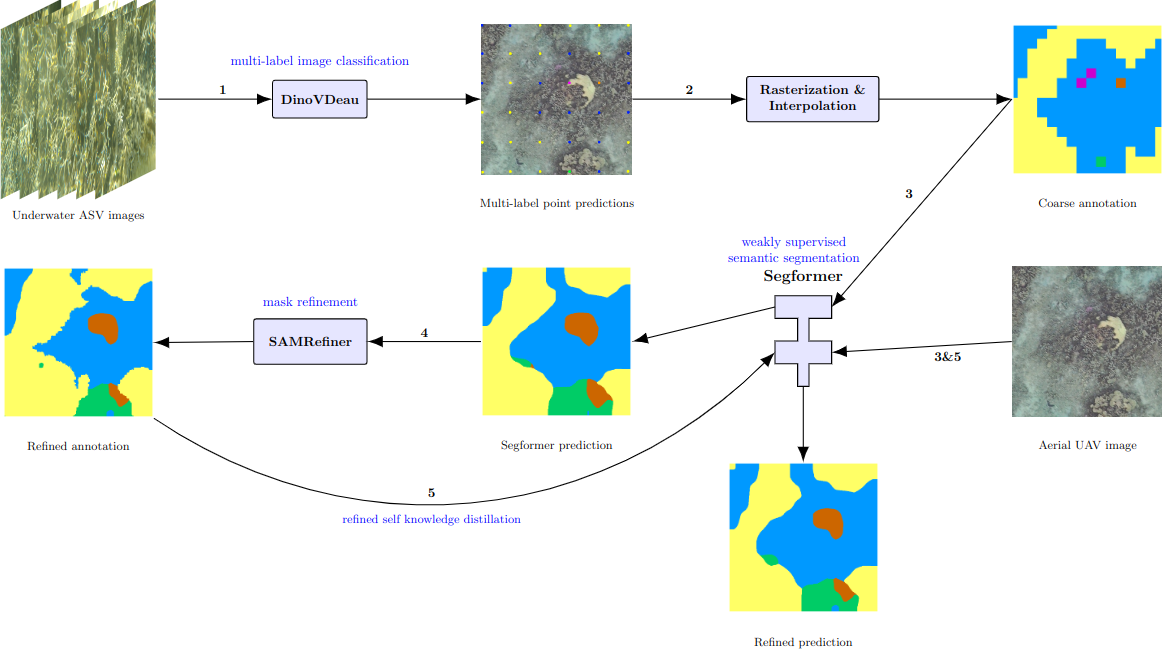}
    \caption{Workflow of the proposed weakly supervised semantic segmentation (WSSS) multi-scale coral reef segmentation approach.
    Underwater images are classified using \textit{DinoVDeau} model (step 1), generating spatialized predictions that are interpolated (arrow 2) to create continuous rasters.
    These rasters are used as coarse annotations for training an Unmanned Aerial Vehicle (UAV)-based segmentation model (step 3) using the SegFormer model.
    Model predictions are then refined using SAMRefiner (step 4) and used to retrain Segformer (step 5), improving segmentation accuracy.}
    \label{fig:schema}
\end{figure}

In this study, we introduce a multi-scale WSSS approach that bridges the gap between fine-scale coral reef monitoring based on underwater images and large-scale mapping based on aerial images. 
The workflow of the proposed method is illustrated in Figure \ref{fig:schema}.
Instead of manually annotating UAV images, we propose using classification-based probability maps derived from underwater images to generate segmentation annotations for aerial imagery. 
Our method spatially interpolates ASV-derived class predictions, producing rough segmentation masks that serve as training labels for SegFormer, a transformer-based semantic segmentation model \cite{NEURIPS2021_64f1f27b}.
UAV segmentation masks are then refined using the SAMRefiner algorithm, a universal and efficient approach that adapts Segment Anything Model (SAM) to the mask refinement task \cite{lin2025samrefiner, kirillov2023segment}. 
The refined masks are then used to retrain the UAV model, improving final segmentation accuracy.

This approach: 
\begin{itemize} 
    \item Transfers fine-scale information from ASV images to UAV segmentation models, reducing the need for manual labelling to the simpler task of classifying underwater images.
    \item Improves the spatial representation of coral morphotypes and habitats in aerial imagery, by passing from image classification to semantic segmentation.
    \item Enables large-scale coral mapping with high-resolution detail, supporting scalable conservation monitoring. 
\end{itemize}

A preliminary version of this work appeared in \citep{contini2025point}.
The present manuscript extends that earlier work in several ways. 
In particular, we introduce methodological improvements to the weakly supervised segmentation framework, including an extended formulation and the addition of a stronger segmentation model. 
The drone-based processing pipeline has also been expanded and the experimental evaluation has been enriched with a new case study on an unseen reef area. 
Furthermore, several sections of the manuscript have been substantially extended, including an expanded discussion of ecological implications, methodological limitations and long-term reef monitoring applications.


\section{Methods} 
\label{sec_matandmet}
This study presents a multi-scale method for generating fine-scale segmentation rasters of coral morphotypes and habitats on UAV imagery. 
The approach relies on collecting and automatically classifying dense underwater images over a portion of the study area to provide fine-scale details for aerial model training.

First, a deep learning model (the teacher model) is used to classify coral morphotypes in ASV images, collected in the Indian Ocean (see \cite{contini2025seatizen} and below for more detailed informations). 
These predictions are then linearly interpolated to generate continuous probability maps across the surveyed area.
To obtain high-resolution raster, a high density of images is required in the surveyed zone. 
For this reason, an ASV is ideal as it allows transects to be spaced less than one metre apart, ensuring complete spatial coverage of the area. 
This enables the generation of reliable, fine-grained rasters that represent the distribution of coral morphotypes.
Since ASV images are geolocated, these rasters can be aligned with aerial images, serving as weak segmentation annotations for training a second model (student) on UAV imagery. 
Finally, the student model predictions are corrected through a mask refinement model, improving the quality of the segmentation masks and enabling the retraining of the model for better accuracy.
To sum up, this new methodology follows the following key steps:

\begin{enumerate}
    \item Underwater data acquisition and image processing: collect high-density geolocated ASV images and classify them using a deep learning teacher model (e.g., as in \cite{contini2025seatizen} and \cite{contini2025underwater}).
    \item Interpolation \& rasterization: create continuous spatial rasters from image classification outputs of the teacher model.
    \item Aerial data acquisition and image processing: build a high-resolution aerial orthophoto from UAV images, ensuring map overlay between ASV spatialized annotation rasters and UAV orthophoto \cite{contini_2025_uav_troudeau, contini_2025_uav_stleu}.
    \item Aerial segmentation dataset construction: split fine-scale annotation rasters and aerial orthophoto into tiles, giving more weight to rare classes.
    \item Aerial model training: use the generated annotations to train a student model for UAV-based coral and habitat classification.
    \item Mask refinement and model retraining: refine student model predictions and retrain a new model to improve segmentation accuracy (see Figure \ref{fig:schema}).
\end{enumerate}

We provide the methodological details of each step in the following subsections. 

\subsection{Underwater data acquisition and image processing}
\label{sec_underwater}

\begin{figure}[ht]
    \centering
    \begin{subfigure}[t]{0.45\textwidth}
        \centering
        \includegraphics[width=\textwidth]{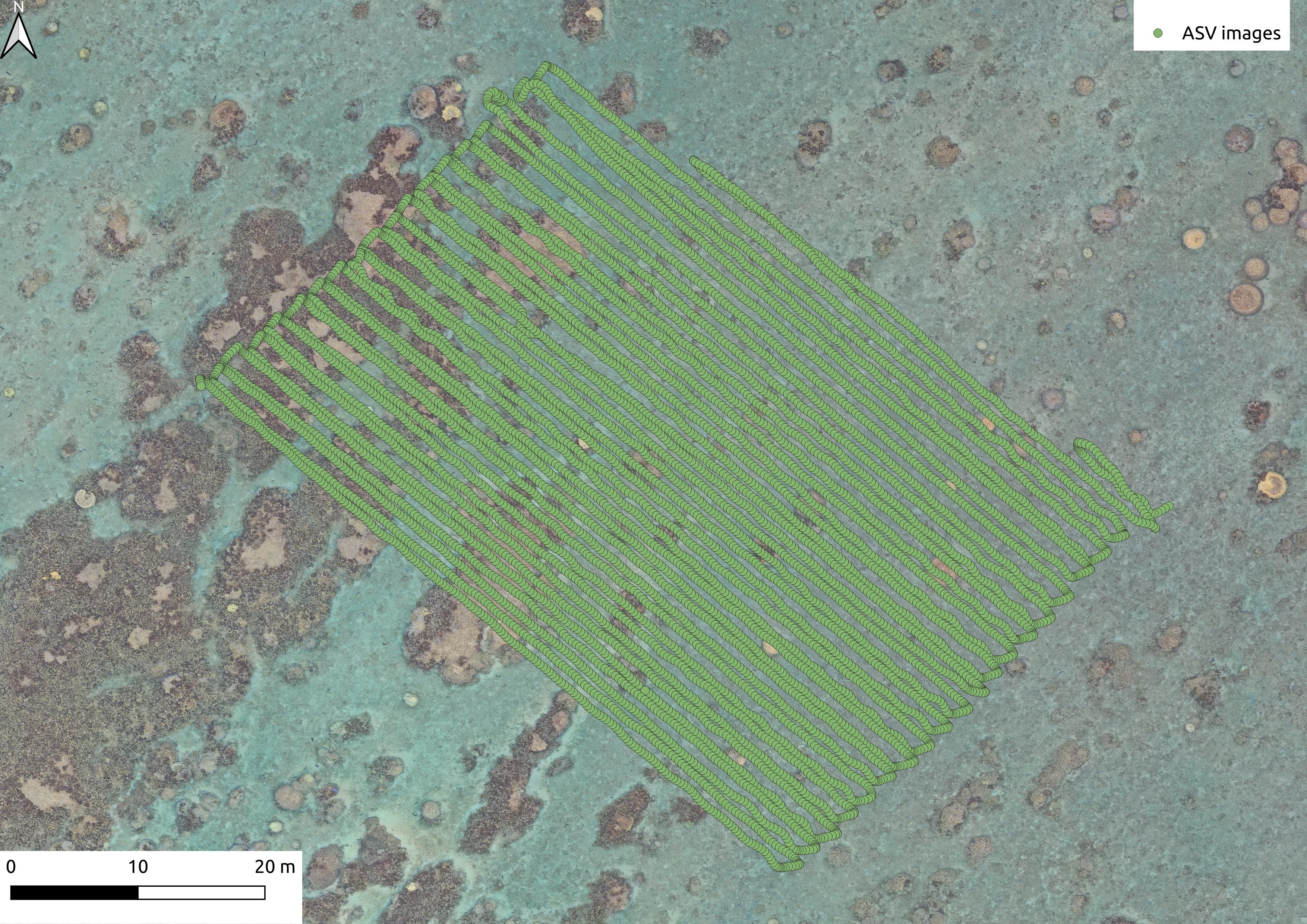}
        \caption{}
        \label{fig:ASV_gps}
    \end{subfigure}
    \hfill
    \begin{subfigure}[t]{0.45\textwidth}
        \centering
        \includegraphics[width=\textwidth]{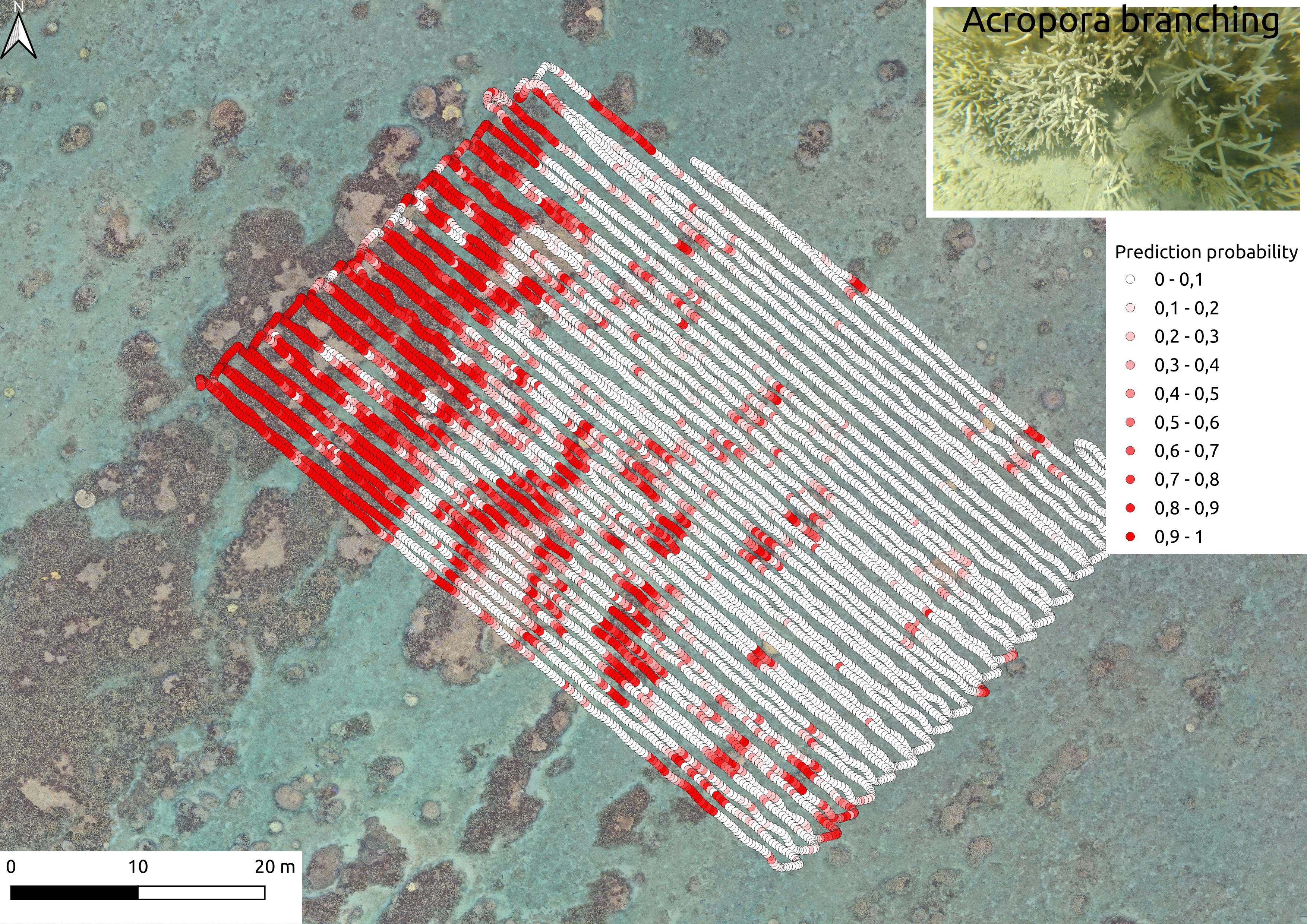}
        \caption{}
        \label{fig:ASV_predictions}
    \end{subfigure}

    \vskip\baselineskip 
    
    \centering

    \begin{subfigure}[t]{0.45\textwidth}
        \centering
        \includegraphics[width=\textwidth]{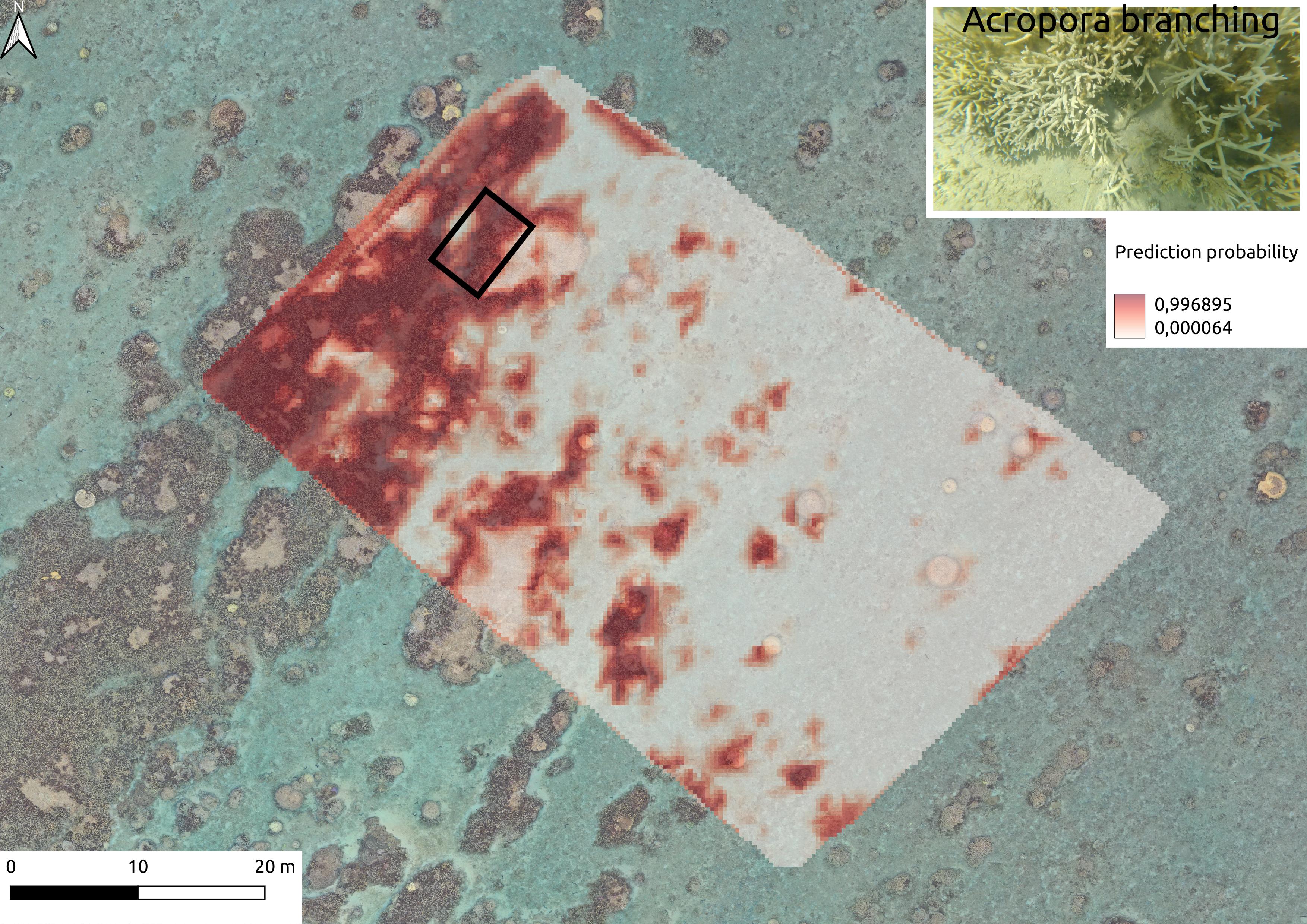}
        \caption{}
        \label{fig:ASV_raster}
    \end{subfigure}
    \hfill
    \begin{subfigure}[t]{0.45\textwidth}
        \centering
        \includegraphics[width=\textwidth]{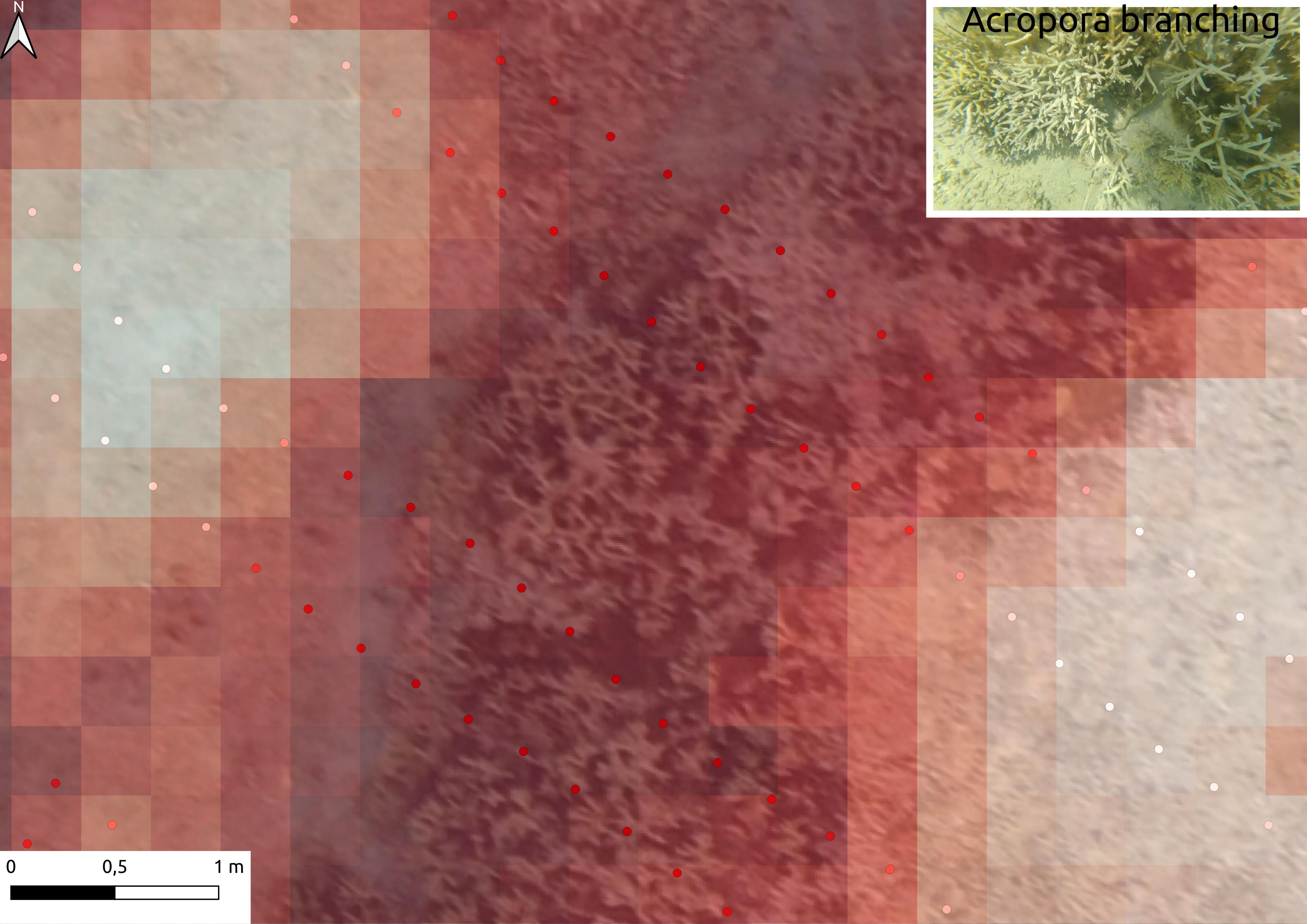}
        \caption{}
        \label{fig:ASV_zoom}
    \end{subfigure}
    \caption{Example of the rasterization process on an ASV data collection event in the \textit{Trou d'eau} lagoon in Reunion Island: 
    (a) GPS coordinates of ASV images along the session.
    The high density of images allows for fine-scale spatial interpolation. 
    (b) Pointwise predictions of $\AcroporeB$ presence in each ASV image. 
    (c) Interpolated raster representing the probability of $\AcroporeB$ presence at each grid point, with the black rectangle indicating the area shown in Figure \ref{fig:ASV_zoom}.
    (d) Zoomed-in view of the raster and corresponding underwater predictions within the black rectangle from Figure \ref{fig:ASV_raster}.
    The raster provides a continuous spatial representation of coral morphotype distributions, enabling the generation of rough segmentation masks that can be used for UAV model training.}
    \label{fig:ASV_rasterization}
\end{figure}

Underwater images were collected using an ASV equipped with a \textit{GoPro Hero 8} camera and a differential GPS \textit{Emlid} Reach M2 mounted on a waterproof case. 
The version of the ASV builds on a previous version developed in \cite{gogendeau2025autonomous}. 
Six ASV missions were conducted along predefined transects, spaced between 0.5 and 1 meter apart (Figure \ref{fig:ASV_rasterization}) in the \textit{Trou d'eau} lagoon on Réunion Island.

Each ASV image (frame extracted from raw videos) was enriched with metadata, including latitude, longitude and attitude metadata (roll, pitch and yaw angles), so that the images could be georeferenced and orientated correctly in the survey area (Figure \ref{fig:ASV_gps}).
More information on time synchronisation between the camera and the GPS clocks and metadata correction can be found in \cite{contini2025underwater}.
To obtain fine-scale predictions of coral morphotypes and habitat types, the deep learning model \textit{DinoVDeau} \cite{DinoVdeau_teacher} trained on the open source dataset \href{https://zenodo.org/records/12819157}{Seatizen Atlas image dataset} \cite{matteo_contini_2024_12819157} is used as a teacher model.
The model outputs probability scores for classes described in \cite{contini2025seatizen}, representing the likelihood of their presence in a given image (in Figure \ref{fig:ASV_predictions}, predictions for the $\AcroporeB$ class are shown on an ASV session). 
Since image classification identifies whether a class is present in an image but does not generate segmentation masks
of the class's exact location, no information about their spatial coverage or location within the frame is provided.

More details about the training and evaluation of the teacher model can be found in Appendix C of \cite{contini2025underwater} and the model itself is publicly available at \cite{DinoVdeau_teacher}.
The teacher model was applied to six ASV sessions conducted in the \textit{Trou d'eau} lagoon, Reunion Island, generating predictions over a total surveyed area  \SI{16904}{\meter\squared} (step 1 in Figure \ref{fig:schema}
).
The ASV sessions used in this study, along with the corresponding preprocessing and inference pipeline, are publicly available and detailed instructions for downloading the data and reproducing the results are provided in Section \ref{sec_data}.

\subsection{Interpolation \& rasterization}
\label{sec_interpolation}
In order to train a semantic segmentation model on UAV images, the ASV pointwise predictions need to be transformed into continuous spatial rasters (step 2 in Figure \ref{fig:schema}).

\begin{figure}[ht]
    \centering
    \includegraphics[width=0.4\linewidth]{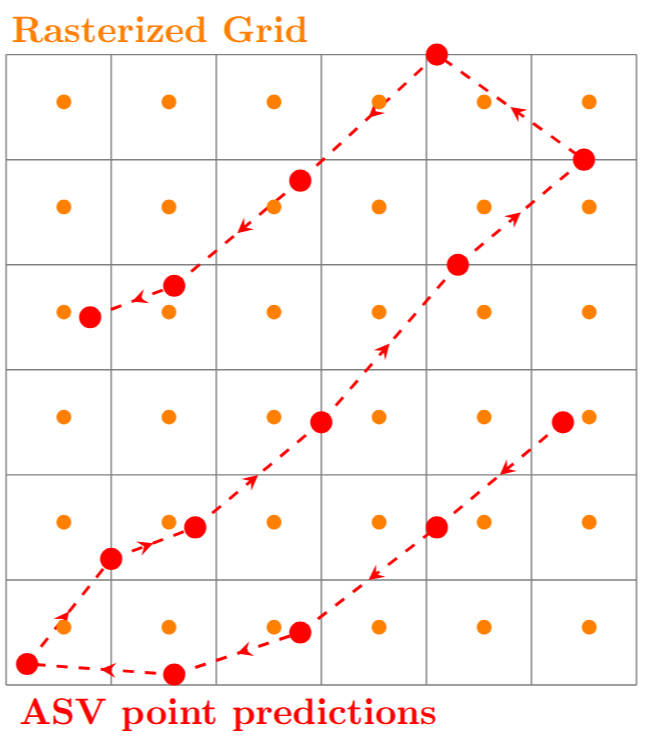}
    \caption{Illustration of the rasterization process used to convert ASV-derived point predictions into a continuous grid. 
    Red dots represent ASV image locations along the survey path, with arrows indicating the acquisition sequence.
    Orange dots mark the centres of the raster grid cells.
    The grid spacing is set to the median distance between two consecutive ASV points, typically ranging from \SI{0.3}{\metre} to \SI{0.35}{\metre}, to ensure sufficient resolution for spatial interpolation.}
    \label{fig:grid}
\end{figure}

To ensure a fine spatial resolution, the grid spacing for rasterization was set to the median distance between two consecutive ASV frames along the survey (Figure \ref{fig:grid}).
Depending on the survey, this distance varied between \SI{0.3}{\metre} and \SI{0.35}{\metre}.
A uniform grid was then created over the survey area, with each grid point representing a pixel in the final raster.
Raster generation was performed using linear interpolation.
We also investigated the impact of the interpolation strategy used to generate the ASV rasters by comparing linear, spline-based and higher-order methods. 
In practice, we observed no significant differences in the final segmentation performance. 
This behaviour is explained by the strong scale mismatch between ASV data (\SI{0.3}{\metre} to \SI{0.35}{\metre} resolution) and UAV imagery (\SI{0.9}{\centi\meter} to \SI{1.6}{\centi\meter} resolution), which requires substantial upsampling of the interpolated rasters and limits the influence of the interpolation method on annotation details. 
For this reason, we adopt linear interpolation, which is simple and computationally efficient, while providing performance comparable to more complex alternatives in our setting.

These rasters are aligned with UAV orthophotos to provide approximate segmentation masks for UAV model training.
Figure \ref{fig:ASV_raster} shows the raster generated from ASV predictions for the $\AcroporeB$ class and in Figure \ref{fig:ASV_zoom} a zoomed-in view of the raster and the corresponding underwater prediction points.

For each ASV data collection event and for each class predicted by the teacher model, a probability raster was generated, representing the likelihood of its presence at every grid point.

\subsection{Aerial data acquisition and image processing}
\label{sec_aerial_data}

Aerial drone images are taken with a \texttt{DJI Mavic 2 Pro} drone. 
A high-resolution UAV orthophoto is constructed by mosaicking individual UAV images. 
Since the drone is not equipped with a differential GPS, once images were taken and the Structure from Motion (SfM) model was built, the orthophoto positioning is refined by collecting ground control points (GCPs) on the surveyed area using a differential GPS.

Two missions were carried out in the lagoon of Réunion Island: one in the \textit{Saint-Leu} lagoon \citep{contini_2025_uav_stleu} and the other in the \textit{Trou d'eau} lagoon \citep{contini_2025_uav_troudeau}, measuring \SI{204748}{\meter\squared} and \SI{189682}{\meter\squared} respectively.
The ground sampling distance (GSD) of the resulting orthophotos was \SI{0.9}{\centi\meter} for \textit{Trou d'eau} and \SI{1.6}{\centi\meter} for \textit{Saint-Leu}, reflecting differences in flight altitude and mission setup.
The model is trained on the \textit{Trou d'eau} lagoon and tested on the \textit{Saint-Leu} lagoon, in order to evaluate its ability to generalize across geographically distinct sites without site-specific tuning.

To ensure spatial consistency between ASV-derived rasters and UAV orthophotos, both datasets are georeferenced using differential GPS measurements and aligned within a common coordinate system. 
A detailed description of the georeferencing, alignment procedure and associated accuracy assessment is provided in Sections 2.1–2.3 of \cite{contini2025underwater}. 
The UAV orthophotos used in this study, along with the corresponding preprocessing workflow, are publicly available and detailed instructions for accessing the data and reproducing the experiments are provided in Section \ref{sec_data}.

\subsection{Aerial segmentation dataset}
\label{sec_aerial_dataset}

\subsubsection{Classes selection}
\label{sec_classes_selection}
\begin{figure}[ht]
    \centering

    \begin{subfigure}[t]{0.45\textwidth}
        \includegraphics[width=\textwidth]{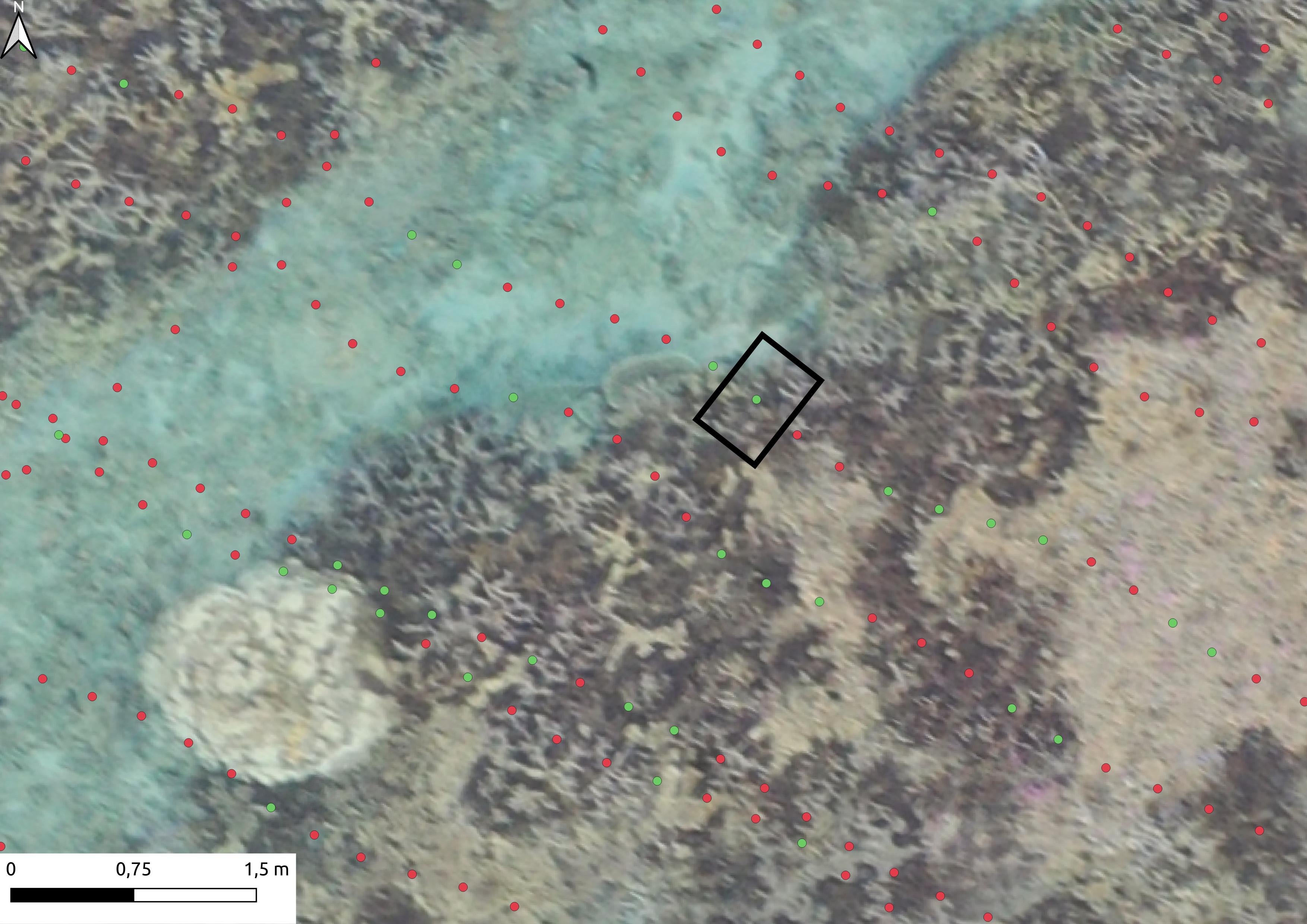}
        \caption{Aerial view of a reef zone with predicted $\AlgaeA$ presence.
         Green dots indicate ASV image locations where the underwater \textit{DinoVdeau} model predicted the presence of $\AlgaeA$; red dots indicate absence.
         The black rectangle highlights the location of the underwater image shown in Figure \ref{fig:supp_unreliable_classes_algae_underwater}, where a dense algal cover is confirmed.}
         \label{fig:supp_unreliable_classes_algae}
    \end{subfigure}
    \hfill
    \begin{subfigure}[t]{0.45\textwidth}
        \includegraphics[width=\textwidth]{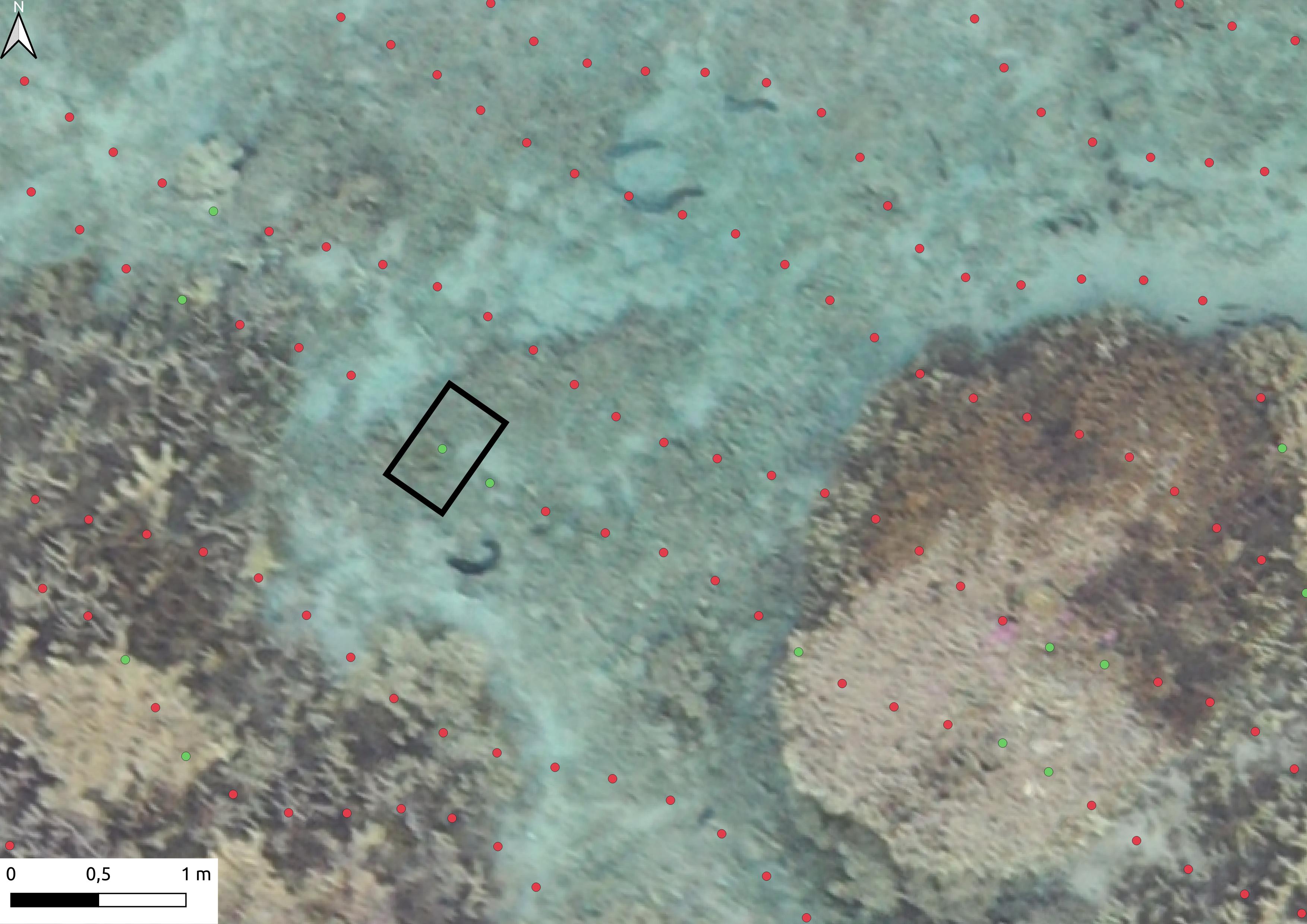}
        \caption{Aerial view of a reef zone with predicted $\AcroporeD$ presence.
        Green dots indicate ASV image locations where the underwater \textit{DinoVdeau} model predicted the presence of $\AcroporeD$; red dots indicate absence.
        The black rectangle highlights the location of the underwater image shown in Figure \ref{fig:supp_unreliable_classes_acropore_underwater}, where a small colony of $\AcroporeD$ is confirmed.}
        \label{fig:supp_unreliable_classes_acropore}
    \end{subfigure}

    \vskip 1em

    \begin{subfigure}[t]{0.45\textwidth}
        \includegraphics[width=\textwidth]{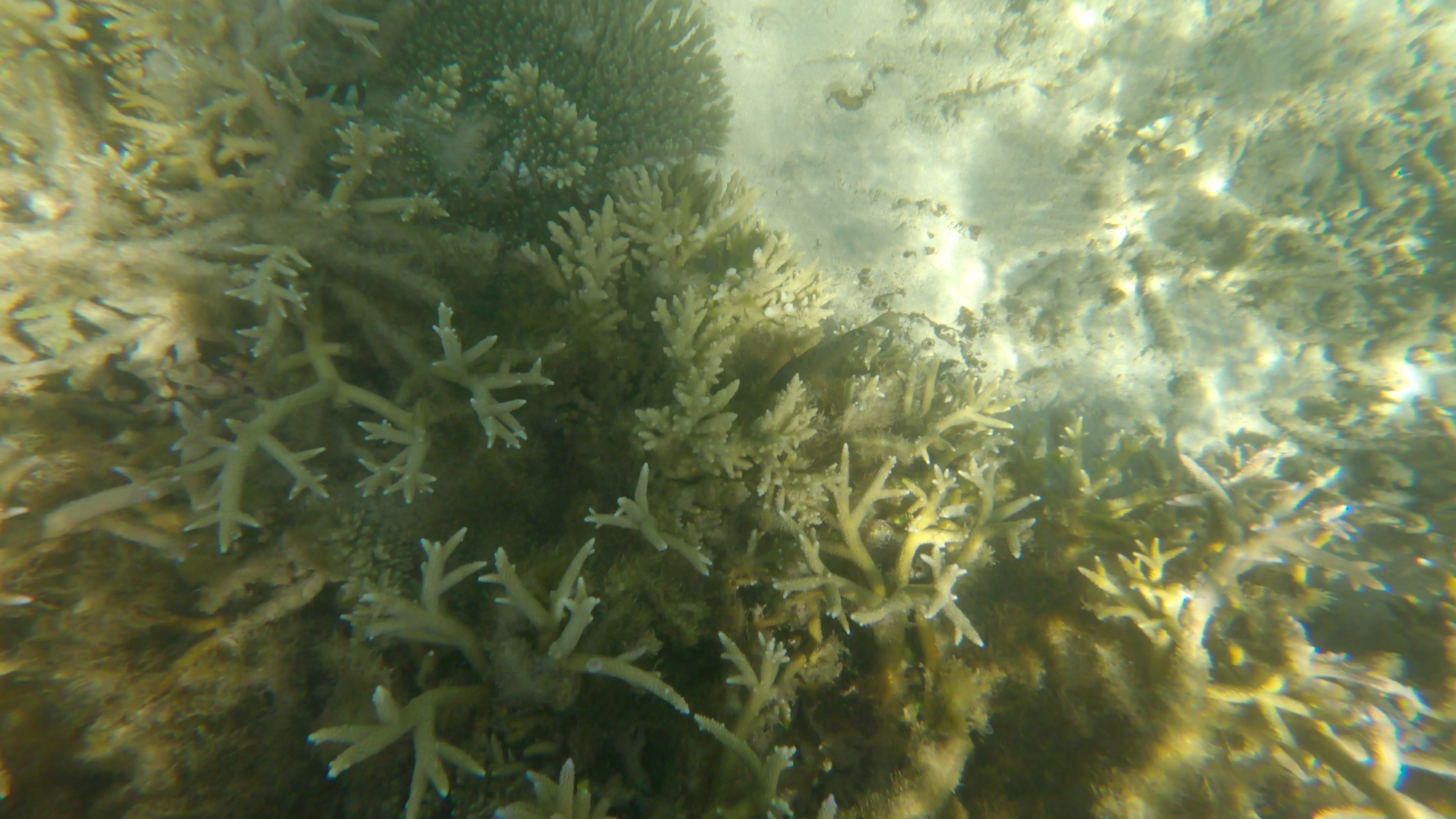}
        \caption{Underwater image taken by an ASV camera showing a dense algal assemblage, confirming \textit{DinoVdeau} prediction at this location.}
        \label{fig:supp_unreliable_classes_algae_underwater}
    \end{subfigure}
    \hfill
    \begin{subfigure}[t]{0.45\textwidth}
        \includegraphics[width=\textwidth]{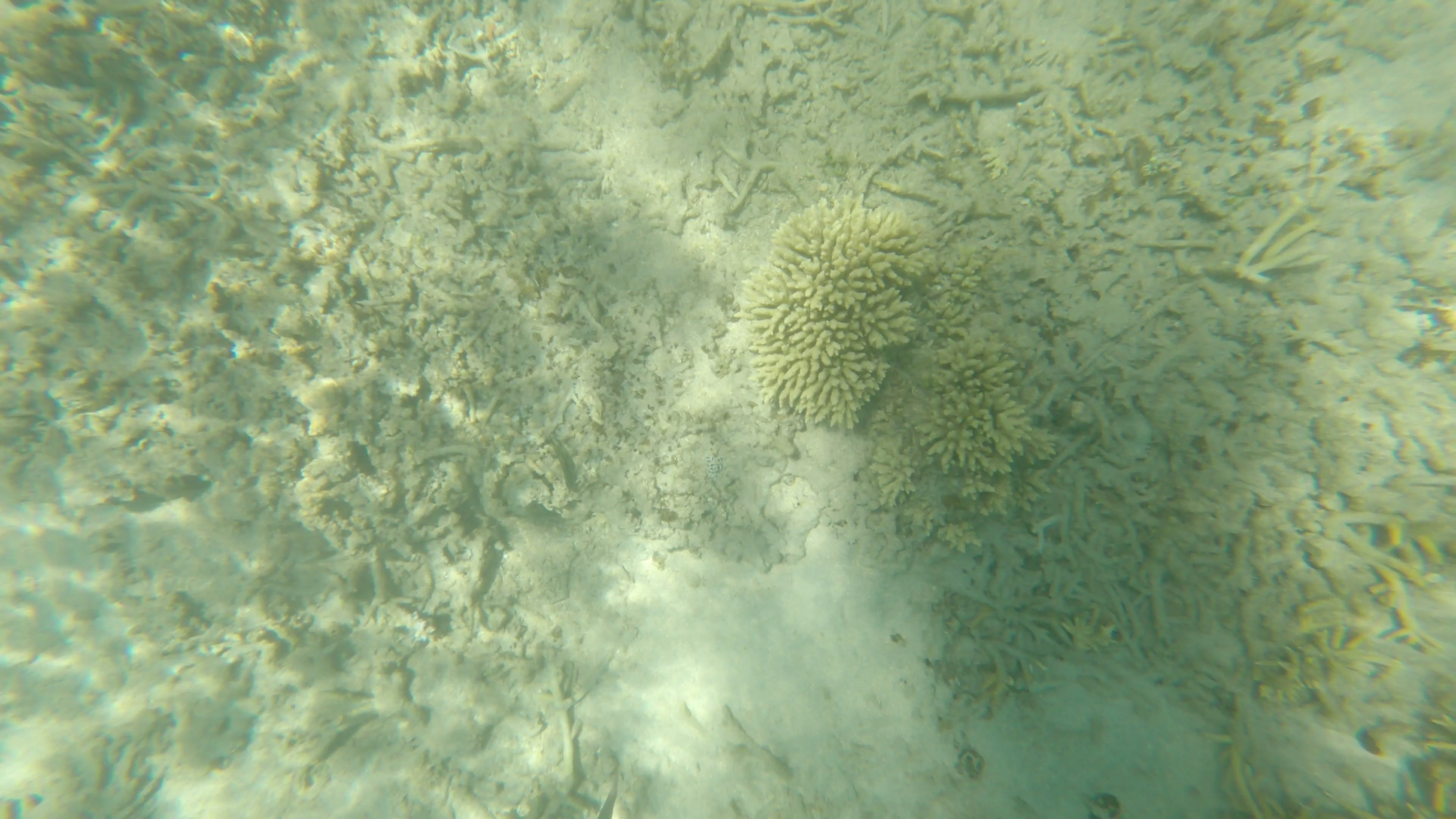}
        \caption{Underwater image showing a small colony of $\AcroporeD$, which is ecologically relevant but not distinguishable in the corresponding aerial orthophoto due to its small size and low contrast.}
        \label{fig:supp_unreliable_classes_acropore_underwater}
    \end{subfigure}

    \caption{Comparison of aerial orthophotos (top) and corresponding underwater images (bottom) for two example locations.
    These examples illustrate how some benthic classes, such as $\AlgaeA$ (left) and small coral morphotypes like $\AcroporeD$ (right), can be clearly identified in underwater imagery but remain indistinct in aerial views due to resolution and colour similarity limitations.}
    \label{fig:supp_unreliable_classes}
\end{figure}

Some classes used in the underwater teacher model cannot be confidently identified in UAV orthophotos due to their size ($\Algae$ classes, small coral colonies like $\AcroporeD$) colour similarity ($\Dead$ and $\Rubble$) or simply lack of resolution in the aerial images.
Figure \ref{fig:supp_unreliable_classes} shows two illustrative examples comparing aerial and underwater views of the same locations, highlighting this limitation.

Figure \ref{fig:supp_unreliable_classes_algae} and Figure \ref{fig:supp_unreliable_classes_acropore} show two aerial orthophotos with green and red dots indicating the presence or absence of $\AlgaeA$ and $\AcroporeD$ predicted by the underwater \textit{DinoVdeau} model on the corresponding ASV images.
The black rectangles highlight the locations of the underwater images shown in Figure \ref{fig:supp_unreliable_classes_algae_underwater} and Figure \ref{fig:supp_unreliable_classes_acropore_underwater}, where the presence of $\AlgaeA$ and $\AcroporeD$ is confirmed.
These examples illustrate how some benthic classes, such as $\AlgaeA$ (left) and small coral morphotypes like $\AcroporeD$ (right), can be clearly identified in underwater imagery but remain indistinct in aerial views due to resolution and colour similarity limitations.
To simplify the segmentation task, we retained only classes that are clearly distinguishable in aerial images.

\begin{figure}[ht]
    \centering
    \includegraphics[width=0.8\textwidth]{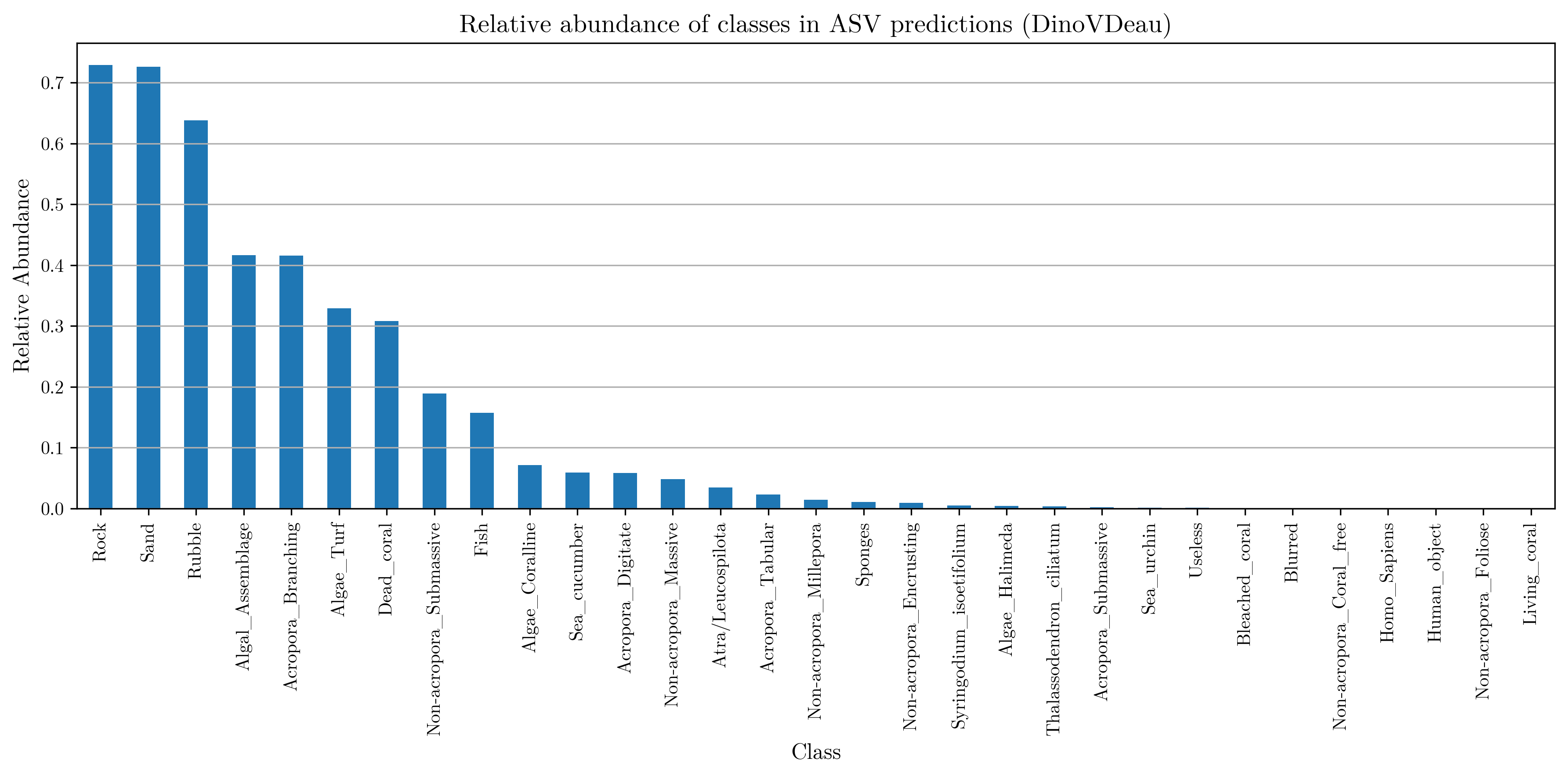}
    \caption{Relative abundance of benthic classes predicted in the underwater imagery by \textit{DinoVdeau} model across the six training ASV sessions.}
    \label{fig:relative_abundance}
\end{figure}

Other classes, although ecologically important and distinguishable on UAV orthophotos like seagrass species $\Syringodium$ and $\Thalassodendron$ were excluded.
This decision was based on the limited number of presence predictions available from the underwater model, which would have increased label noise during training.
Figure \ref{fig:relative_abundance} presents the relative abundance of each of the 31 benthic classes predicted in the underwater imagery by \textit{DinoVdeau} model, on the six ASV sessions retained for this study.

Hard-substrate classes like \textit{Rock}, \textit{Sand} and \textit{Rubble} dominate the distribution but only \textit{Sand} was retained in this study. 
\textit{Rock} and \textit{Rubble} were excluded because they appear too similar to \textit{Sand} in UAV imagery at the given resolution. 
Additionally, their frequent co-occurrence in underwater predictions would have introduced significant label noise during the generation of coarse annotations.
It is also important to note that the relative abundance values shown in Figure \ref{fig:relative_abundance} do not sum to 1. 
This is because the underwater model \textit{DinoVdeau} is a multi-label image classification model, allowing multiple classes to be predicted for a single image, reflecting the natural co-occurrence of benthic features in reef environments.

Thus, for the purpose of a proof of concept, we selected the following subset of classes, which exhibit clear visual patterns in aerial images:

\begin{enumerate}

\begin{multicols}{2}
    \item $\AcroporeB$
    \item $\AcroporeT$
    \item $\NoAcroporeM$
    \item \textit{Other Corals}
    \item $\Sand$ 
\end{multicols}
         
\end{enumerate} 

The first two classes are common coral morphotypes of the genus \textit{Acropora} that can be found in Réunion Island's lagoons.
$\AcroporeB$ is a branching coral easily distinguishable from aerial images due to its particular spectral signature (dark brown) and is often found in large colonies. 
$\AcroporeT$ is a tabular coral a little bit more difficult to identify on aerial images but still distinguishable due to the particular layer shape of its colonies.
The third class $\NoAcroporeM$ is a stony coral with a light spectral signature, often found alone surrounded by sand.
The fourth class \textit{Other Corals} encompasses many different species of corals that can range from small colourful colonies of the genus \textit{Pocillopore} to large stony corals that are darker in colour than $\NoAcroporeM$ corals.

\subsubsection{Annotation generation}
\label{sec_annotation_generation}

In the semantic segmentation task, a model is trained to predict a class label for each pixel in an image. 
To achieve this, a set of images must be labelled so that each pixel corresponds to one and only one class. 
Let $C$ be the set of classes, $I$ be the set of images and $N$ the cardinality of $C$.

In our case, for an image $i \in I$, we have $N$ continuous rasters representing the probability of presence for each class $c \in C$. 
Thus, we need a function that, for each pixel, maps these probabilities to a class label. 
A straightforward approach is to assign to each pixel the label of the class with the highest probability of presence.

\begin{figure}[ht]
    \centering
    \begin{subfigure}[t]{0.45\textwidth}
        \centering
        \includegraphics[width=\textwidth]{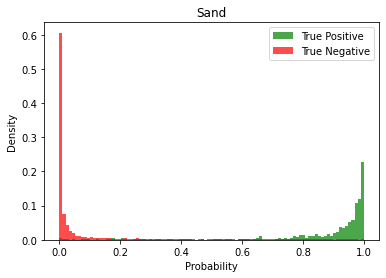}
        \caption{}
        \label{fig:hist_sand}
    \end{subfigure}
    \hfill
    \begin{subfigure}[t]{0.45\textwidth}
        \centering
        \includegraphics[width=\textwidth]{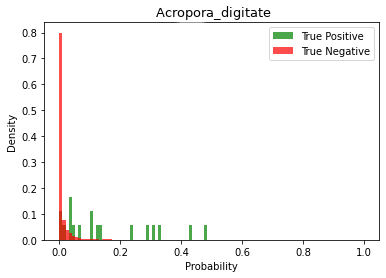}
        \caption{}
        \label{fig:hist_ad}
    \end{subfigure}

    \caption{Probability distribution of true positives and true negatives for two classes in the underwater image classification teacher model.
    (a) $\Sand$ class is a common class with high probabilities of presence when correctly detected and low probabilities when absent.
    (b) $\AcroporeD$ class is a rare class with a low probability of presence, even when correctly detected.}
    
    \label{fig:hist}
\end{figure}

However, this approach has limitations. 
Since the underwater teacher model is a multilabel classifier (trained with the binary cross-entropy loss function), the predicted presence/absence probabilities of each class are independent and not necessarily calibrated in the same way. 
As an illustration, Figure \ref{fig:hist} shows the distribution of the predictions of true positives and true negatives for two classes.

For common classes such as $\Sand$, the teacher model tends to assign very high probabilities to true positives and low values to true negatives, resulting in well-separated bimodal distributions (Figure \ref{fig:hist_sand}). 
In contrast, for rare classes such as $\AcroporeD$, the model shows a lower confidence, with true positive predictions spread more uniformly throughout the probability range (Figure \ref{fig:hist_ad}). 
This leads to a systematic underestimation of rare classes when using naïve maximum-probability labelling.

To address this, we implement a class-specific quantile normalisation strategy. 
Following the principles outlined in \citet{zhao2020quantile}, we normalise the predicted probability maps for each class independently using their empirical percentiles. 
Specifically, for each class $c$, we aggregate all predicted probabilities generated by the teacher model across all ASV sessions and compute the 1st and 99th percentiles. 
We then linearly scaled the per-pixel probability values within this range. 
This ensures that for each class, the normalised values reflect relative confidence within that class's distribution, rather than across the entire set.

Formally, let $p_c(x, y)$ denote the predicted probability of presence for class $c$ at pixel location $(x, y)$ in the ASV-derived raster.
Let $q_{1}(c)$ and $q_{99}(c)$ be the empirical 1\textsuperscript{st} and 99\textsuperscript{th} percentile values of the predicted probability distribution for class $c$ across all sessions. 
The normalised probability $\hat{p}_c(x, y)$ is computed as:

\begin{equation} 
    \hat{p}_c(x, y) = \text{clip}\left( \frac{p_c(x, y) - q_{1}(c)}{q_{99}(c) - q_{1}(c) + \epsilon},\ 0,\ 1 \right) 
\end{equation}

where $\epsilon$ is a small constant added for numerical stability and the clip function ensures that values remain within the valid probability range. 
This normalisation method improves the naïve maximum selection by taking into account the class-specific prediction dynamics and mitigating the bias caused by imbalanced raw probability distributions. 
The final label assigned to each pixel is then:

\begin{equation} 
    L(x, y) = \arg\max_{c \in C} \hat{p}_c(x, y) 
\end{equation}

This quantile-based normalisation is conceptually similar to class-specific normalisation strategies explored in gene expression data preprocessing, where non-uniform expression distributions across classes can lead to biased analyses if left uncorrected \citep{zhao2020quantile}. 
Applying this principle to semantic segmentation in ecology helps rectify inter-class imbalance, leading to a more equitable representation of rare classes in the resulting annotations.

Finally, since the aerial orthophoto and the ASV annotation rasters do not share the same resolution and since semantic segmentation models require a one-to-one pixel correspondence between the image and annotation, we upsample the ASV annotation rasters to the resolution of the orthophoto by replicating nearest‑neighbour pixel values.

\subsection{Training dataset statistics}

The coarse annotations used for training are generated from a total of 59,433 ASV image frames collected across six sessions. 
These geolocated predictions are spatially interpolated and aligned with UAV orthophotos to produce weak segmentation masks.
The resulting dataset comprises 519 UAV image tiles of size $512 \times 512$ pixels, of which 363 are used for training and 156 for validation. 

The distribution of pixels across classes in the generated coarse annotations is summarized in Table \ref{tab:class_distribution}.
Due to the nature of benthic habitats, the dataset is highly imbalanced, with dominant classes such as sand covering a large proportion of the area.

\begin{table}[h]
\centering
\caption{Pixel distribution of training annotations across classes.}
\label{tab:class_distribution}
\begin{tabular}{lcc}
\hline
Class & Number of pixels & Percentage (\%) \\
\hline
Sand & 89,774,885 & 65.98 \\
Acropora Branching & 36,731,606 & 27.00 \\
Other Corals & 5,634,688 & 4.14 \\
Acropora Tabular & 2,118,214 & 1.56 \\
Non-acropora Massive & 1,793,343 & 1.32 \\
\hline
\end{tabular}
\end{table}

\subsection{Aerial deep learning model (student model)}
\label{subsec_aerial_deep_learning_model}

WSSS provides a framework for training models using approximate labels instead of manually annotated ground truth masks \cite{chan2021comprehensive}.
In our case, the annotations are derived from ASV-based probability maps, which are inherently coarse and may contain noise.
These weak labels, combined with the challenges of UAV-based segmentation (such as variable lighting conditions, water clarity differences and limited spatial resolution that obscures fine coral structures) introduce significant uncertainty into the training process.
As a result, the segmentation model must be robust to label noise and able to generalise to different spatial conditions and environmental contexts.

To address these requirements, we adopted SegFormer, a transformer-based semantic segmentation architecture that provides five model variants with increasing representational capacity. 
In this study, we selected two configurations: the \textit{B0} variant, corresponding to the smallest available model and the \textit{B2} variant, which represents the largest configuration that could be trained with our available computational resources.
This choice allowed us to investigate the effect of model scale while remaining compatible with the hardware constraints of the training setup. 
SegFormer offers several advantages over conventional convolutional neural networks \cite{NEURIPS2021_64f1f27b, ru2022learning}:

\begin{itemize}
\item Hierarchical feature representation: SegFormer processes images at multiple scales, allowing it to capture both global contextual information and fine-grained details. 
This is particularly useful for UAV imagery, where coral morphotypes may have highly variable spatial patterns.
\item Robust to weak annotations: unlike models that rely on precise pixel-wise labels, SegFormer effectively learns from weak labels by employing an efficient self-attention mechanism, improving segmentation accuracy even when trained on approximate masks.
\item Lightweight and efficient: unlike traditional vision transformers, SegFormer does not require heavy computational resources, making it feasible for large-scale aerial image processing.
\end{itemize}

Choosing an appropriate loss function is critical in WSSS, where training annotations are coarse and often affected by noise.
Region-based losses, such as Dice loss, are especially suitable in these scenarios because they evaluate the overlap between predicted and annotated regions as a whole, rather than focussing on individual pixels or precise edges \cite{azad2023loss}.
Unlike pixel-wise losses, which are sensitive to misaligned labels, or boundary-based losses, which require accurate contour annotations, region-based losses are more tolerant of uncertainty in object boundaries, making them a better fit for learning from approximate masks.
Given the nature of this problem, we opted to minimise the region-based Dice loss with an Adam optimiser \cite{kingma2014adam}.

The input and output resolutions are chosen to be \SI{512}{\pixel} $\times$ \SI{512}{\pixel}, with a batch size of 16 images.
The initial learning rate was initially set to $10^{-5}$ and decayed by a factor of 10 whenever the validation loss plateaued for five consecutive epochs.

\subsection{Mask refinement and model retraining}
\label{subsec_mask_refinement}
The final step in our methodology involves refining the initial segmentation masks generated by the student model (step 4 in Figure \ref{fig:schema}).
A common technique in computer vision for improving model performance without additional supervision is Self Knowledge Distillation (SKD), where the model is retrained using its own predictions as pseudo-labels \cite{xu2024survey}.
In this configuration, the same model assumes the roles of both teacher and student, iteratively improving itself by distilling and refining its own previously generated output.

In this work, self-distillation is implemented as a two-stage training strategy. 
First, a SegFormer model is trained using coarse segmentation masks derived from ASV-based interpolated probability maps. 
This initial model is then applied to the UAV imagery to generate dense pixel-wise predictions, which serve as pseudo-labels.
To improve the spatial consistency of these pseudo-labels, we explore two variants: (i) directly retraining the model using its own predictions (standard SKD) and (ii) refining these predictions prior to retraining using a mask refinement model. 
In the latter case, we employ SAMRefiner, a universal and efficient approach that adapts SAM to mask refinement tasks \cite{lin2025samrefiner}.
A second SegFormer model is then trained from scratch using the resulting pseudo-labels (step 5 in Figure \ref{fig:schema}).

\subsection{Test zone and model evaluation}
\label{subsec_test_zone}

Due to the lack of human-made annotations, the deep learning model cannot be evaluated on weakly generated annotations using classic computer vision metrics on the test set.
For this reason, we selected two test zones: one within the \textit{Trou d'eau} lagoon measuring \SI{84}{\meter\squared} and the other in the \textit{Saint-Leu} lagoon measuring \SI{194}{\meter\squared}.

These zones were chosen due to their diverse composition of coral morphotypes, habitats and other marine organisms, representing a challenging environment for model validation \cite{contini_2025_uav_troudeau, contini_2025_uav_stleu}.

\section{Results} 
\label{sec_results_chap4}

\subsection{Evaluation of model variants during pipeline development}
\label{subsec_pipeline_evaluation}
To justify the contribution of each component in the proposed pipeline (illustrated in Figure \ref{fig:schema}), we evaluated the segmentation performance of the \textit{B0} model variant on manually annotated ground-truth data from the test zones in the \textit{Trou d'eau} and \textit{Saint-Leu} lagoons (see Section \ref{subsec_test_zone}).
The following model configurations were tested to isolate the effect of each step in the pipeline:

\begin{itemize}
    \item \textit{Coarse model}: Segformer trained solely on coarse annotations derived from ASV predictions (step 3 in Figure \ref{fig:schema}). 
    No mask refinement or self-distillation is applied.
    
    \item \textit{Coarse refined}: coarse annotations are first refined using the SAMRefiner model. 
    Segformer is then trained on these refined masks.
    No self-distillation is applied.
    
    \item \textit{Coarse self-distilled}: Segformer is initially trained on coarse annotations and then retrained using its own predictions as supervision (self-distillation), without involving SAMRefiner.
    
    \item \textit{Coarse refined + distilled}: full pipeline.
    Segformer is first trained on coarse annotations, whose predictions are then refined using SAMRefiner. 
    A final Segformer is trained via self-distillation on these refined predictions.
    This is the final proposed model (step 5 in Figure \ref{fig:schema}).
\end{itemize}

Table \ref{tab:model_comparison_iou} presents total pixel accuracy, mean IoU and per-class IoU for each model.
For a more detailed definition of these metrics, please refer to \citep{schlosser2024consolidated}.

\begin{table}[ht]
\small
\centering
\caption{Comparison of total accuracy and mean IoU across different model variants, along with IoU per class.}
\label{tab:model_comparison_iou}
\resizebox{\columnwidth}{!}{%
\begin{tabular}{>{\scriptsize}l >{\scriptsize}c >{\scriptsize}c | >{\scriptsize}c >{\scriptsize}c >{\scriptsize}c >{\scriptsize}c >{\scriptsize}c}
    \toprule
    Model & Accuracy & Mean IoU &
    Acropora B. & Acropora T. & Non-acro M. & Other Corals & Sand \\
    \midrule
    \textit{Coarse} & 0.8250 & 0.4625 & 0.5109 & 0.2087 & 0.2877 & \textbf{0.4201} & 0.8854 \\
    \textit{Coarse Refined} & 0.8467 & 0.4658 & 0.4676 & 0.1532 & 0.3721 & 0.4078 & \textbf{0.9285} \\
    \textit{Coarse Self-Distilled} & 0.8300 & 0.4745 & 0.5175 & 0.2114 & 0.3395 & 0.4190 & 0.8852 \\
    \textit{Coarse Refined + Distilled} & \textbf{0.8572} & \textbf{0.5010} & \textbf{0.5282} & \textbf{0.2313} & \textbf{0.4095} & 0.4111 & 0.9249 \\
    \bottomrule
\end{tabular}
}
\end{table}

The results show that although the \textit{Coarse} and the \textit{Coarse refined} model variants achieve a slightly better IoU for \textit{Other Corals} and $\Sand$ classes respectively, only the full pipeline offers a more balanced trade-off between pixel-level accuracy and spatial precision, as indicated by its superior accuracy and mean IoU scores.
This progressive evaluation validates the effectiveness of the proposed weakly supervised multi-scale training framework.  

In the following sections, we focus on the results obtained with the full pipeline model to evaluate the quality and ecological relevance of the segmentation outputs across annotated and unannotated reef zones.

\subsection{Comparison between SegFormer B0 and B2 variants}
\label{subsec_b0_b2_comparison}
To assess the impact of model capacity on segmentation performance, we further compared the final pipeline using the \textit{B0} and \textit{B2} SegFormer variants. 
Both models were trained using the full pipeline configuration (\textit{Coarse refined + distilled}) and evaluated on the same manually annotated test data.

Table \ref{tab:b0_b2_comparison} reports pixel accuracy, mean accuracy and mean IoU for the two variants.

\begin{table}[ht]
\small
\centering
\caption{Performance comparison between SegFormer B0 and B2 variants using the full pipeline.}
\label{tab:b0_b2_comparison}
\begin{tabular}{lccc}
\toprule
Model Variant & Pixel Accuracy & Mean Accuracy & Mean IoU \\
\midrule
SegFormer B0 & 0.8572 & 0.7793 & 0.5010 \\
SegFormer B2 & \textbf{0.8607} & \textbf{0.7871} & \textbf{0.5223} \\
\bottomrule
\end{tabular}
\end{table}

The results show that increasing model capacity from \textit{B0} to \textit{B2} improves performance across all evaluation metrics. 

\subsection{Model performance on annotated reef zones}
\label{subsec_upscaling_evaluation}
We now focus on the performance of the final model (i.e., the \textit{Coarse refined + distilled} variant of Segformer \textit{B2}) on the same annotated test zones used in the previous section, namely the \textit{Trou d'eau} and \textit{Saint-Leu} lagoons (see Section \ref{subsec_test_zone}).
This assessment aims to characterise the spatial accuracy and ecological relevance of the predicted segmentation masks at the UAV scale.
Per-zone and per-class metrics are presented in Table \ref{tab:iou_by_zone}.

\begin{table}[ht]
\small
\centering
\caption{IoU per class, total accuracy and mean IoU by zone.}
\label{tab:iou_by_zone}
\resizebox{\columnwidth}{!}{%
\begin{tabular}{>{\scriptsize}l >{\scriptsize}c >{\scriptsize}c | >{\scriptsize}c >{\scriptsize}c >{\scriptsize}c >{\scriptsize}c >{\scriptsize}c}
    \toprule
    Zone & Accuracy & Mean IoU &
    Acropora B. & Acropora T. & Non-acro M. & Other Corals & Sand \\
    \midrule
    \textit{Trou d'eau} & 0.9297 & 0.5609 & 0.2471 & 0.5531 & 0.5914 & 0.4544 & 0.9586 \\
    \textit{Saint-Leu}  & 0.8016 & 0.5354 & 0.4750 & / & 0.4427 & 0.3150 & 0.9089 \\
    \textbf{Total}      & \textbf{0.8607} & \textbf{0.5223} & \textbf{0.4505} & \textbf{0.4016} & \textbf{0.4757} & \textbf{0.3478} & \textbf{0.9356} \\
    \bottomrule
\end{tabular}
}
\end{table}

\begin{figure}[ht]
    \centering
    \includegraphics[width=0.7\textwidth]{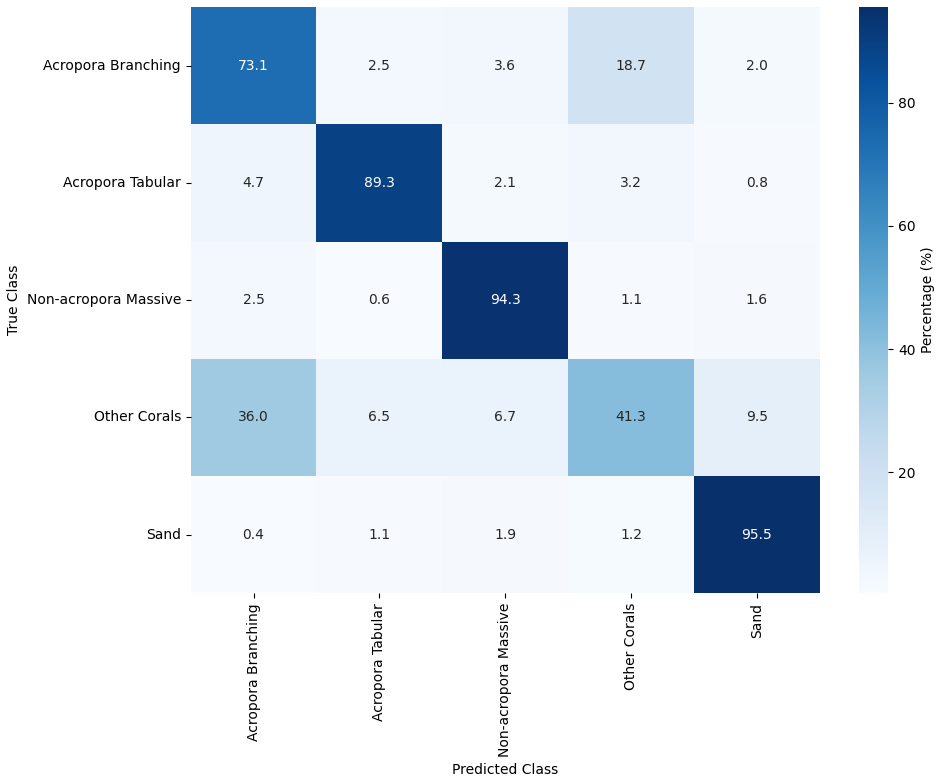}
    \caption{Normalized confusion matrix computed across both test zones (\textit{Trou d'eau} and \textit{Saint-Leu}). 
    Values represent the percentage of true class pixels predicted for each class.}
    \label{fig:confusion_matrix}
\end{figure}

The total pixel accuracy reached 86.07\%, while the mean IoU was 52.23\%.
The confusion matrix, showing the percentage of true class pixels predicted for each class, along with misclassification errors, is presented in Figure \ref{fig:confusion_matrix}.

To assess the statistical reliability of these results, we additionally estimated 95\% confidence intervals for all metrics using a spatial block bootstrap procedure over the annotated test zones. 
Each zone was partitioned into non-overlapping tiles of $128 \times 128$ pixels, resulting in 45 and 50 blocks for the \textit{Trou d'eau} and \textit{Saint-Leu} zones, respectively.
Bootstrap samples were drawn with replacement at the tile level and metrics were recomputed for each resample ($N=1000$). 
Confidence intervals were obtained from the \num{2.5}\textsuperscript{th} and \num{97.5}\textsuperscript{th} percentiles of the resulting distributions.
The estimated performance metrics and their associated confidence intervals are summarized in Table \ref{tab:bootstrap_ci}.

\begin{table}[ht]
\small
\centering
\caption{Performance metrics with 95\% confidence intervals estimated using spatial block bootstrap.}
\label{tab:bootstrap_ci}
\begin{tabular}{lcc}
\hline
\textbf{Metric} & \textbf{Estimate} & \textbf{95\% CI} \\
\hline
Pixel Accuracy & 0.8473 & [0.8019, 0.8898] \\
Mean IoU       & 0.5121 & [0.4280, 0.5737] \\
\hline
\textbf{IoU per class} & & \\
Acropora Branching     & 0.4385 & [0.2588, 0.6077] \\
Acropora Tabular       & 0.4138 & [0.0855, 0.6003] \\
Non-acropora Massive   & 0.4640 & [0.2970, 0.5996] \\
Other Corals           & 0.3160 & [0.2123, 0.4236] \\
Sand                   & 0.9282 & [0.9049, 0.9483] \\
\hline
\end{tabular}
\end{table}

\subsection{Predicted habitat composition across reef zones}
\label{subsec_prediction_masks}

\begin{figure}[ht!]
    \centering
    \begin{subfigure}[t]{0.45\textwidth}
        \centering
        \includegraphics[width=\textwidth]{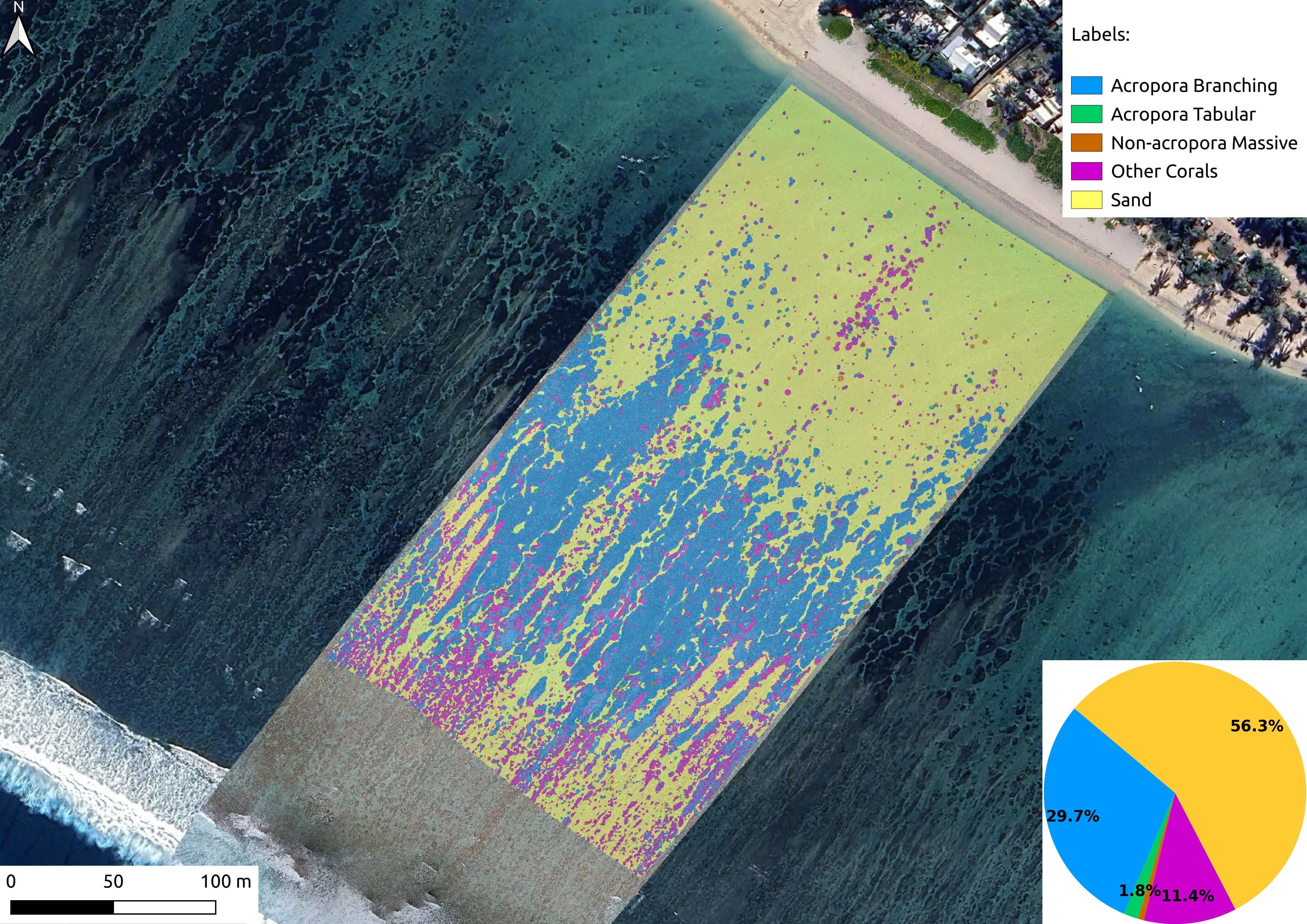}
        \caption{}
        \label{fig:troudeau_prediction}
    \end{subfigure}
    \hfill
    \begin{subfigure}[t]{0.45\textwidth}
        \centering
        \includegraphics[width=\textwidth]{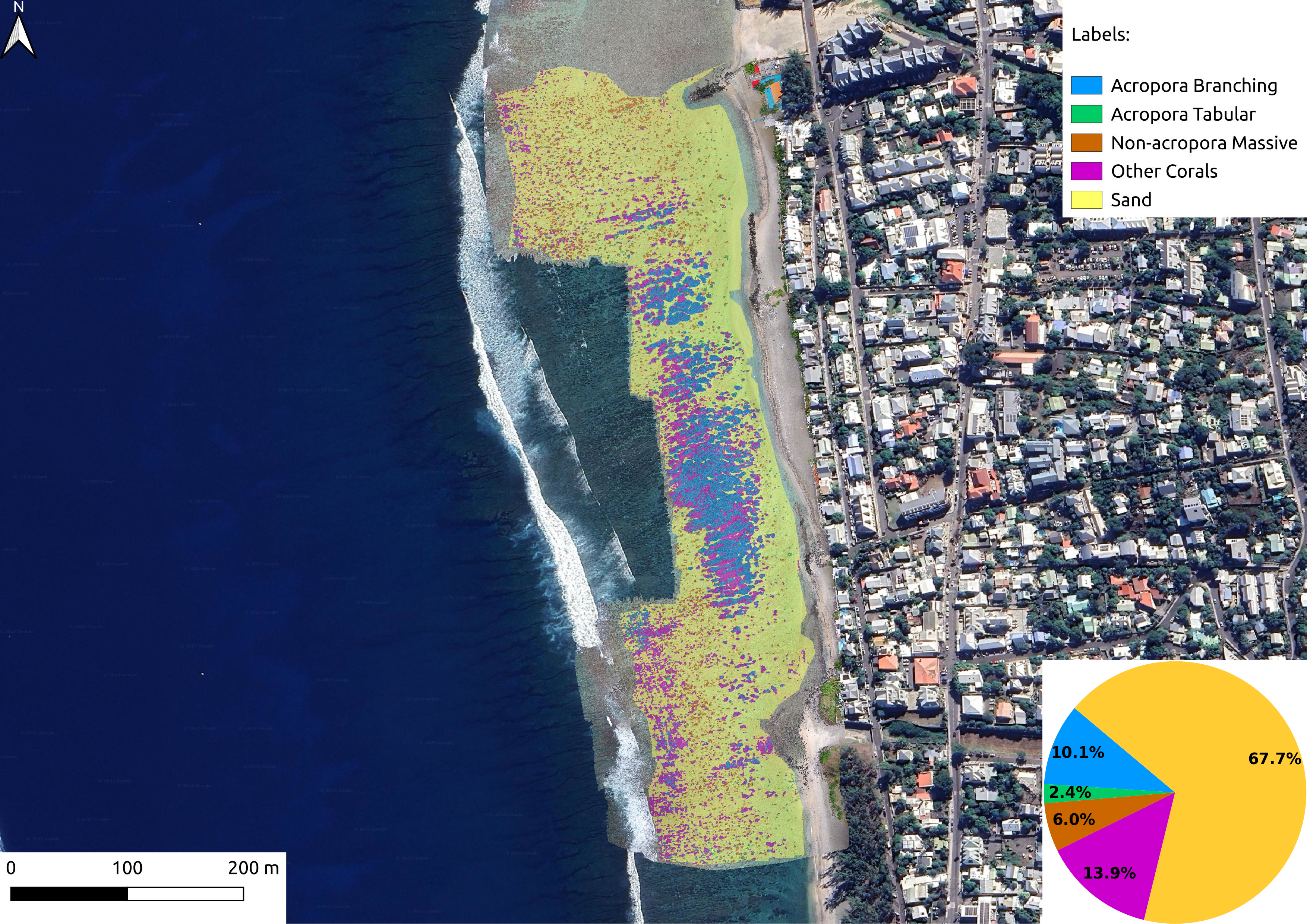}
        \caption{}
        \label{fig:stleu_prediction}
    \end{subfigure}

    \vskip\baselineskip 
    
    \centering

    \begin{subfigure}[t]{0.45\textwidth}
        \centering
        \includegraphics[width=\textwidth]{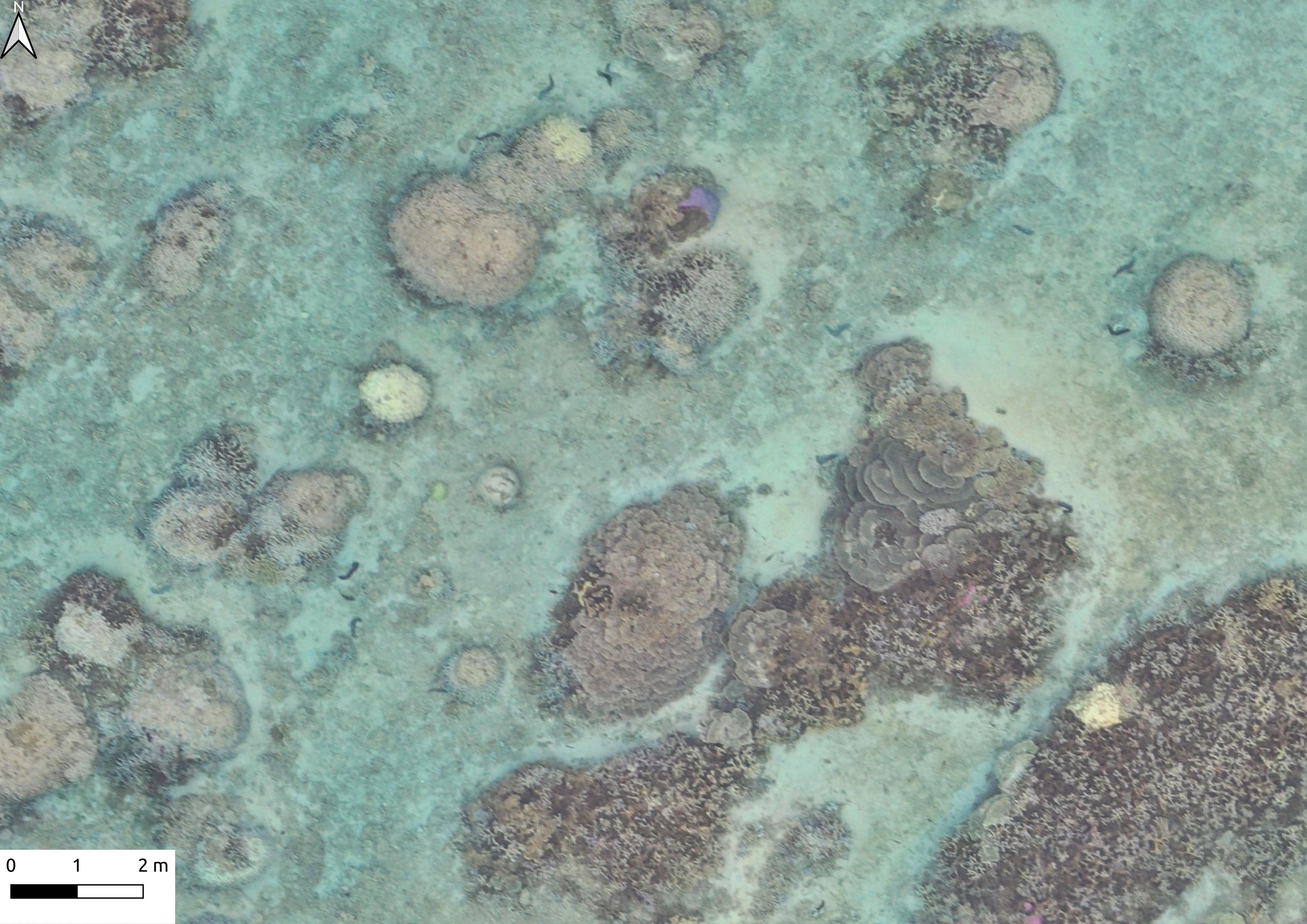}
        \caption{}
        \label{fig:troudeau_ortho}
    \end{subfigure}
    \hfill
    \begin{subfigure}[t]{0.45\textwidth}
        \centering
        \includegraphics[width=\textwidth]{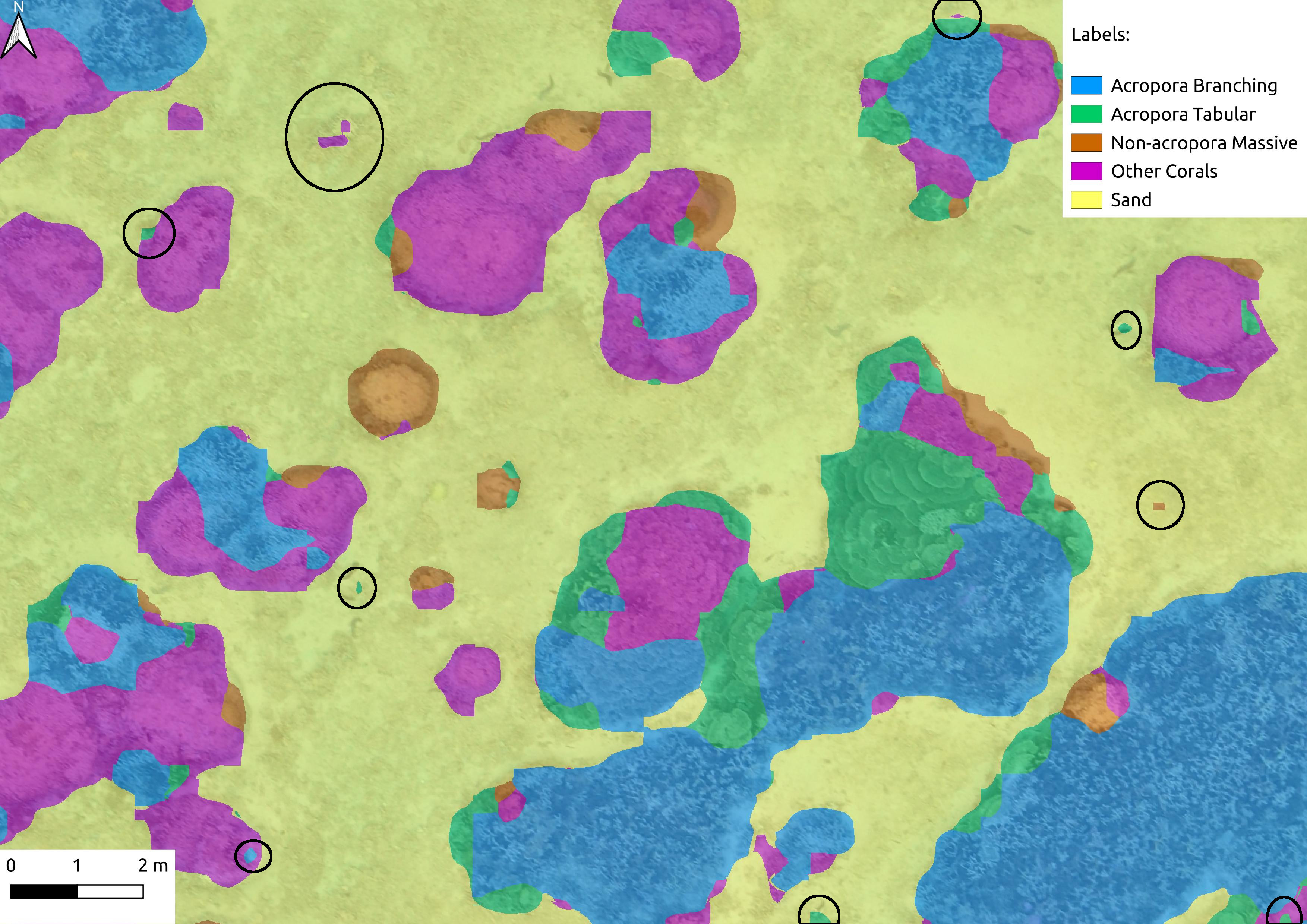}
        \caption{}
        \label{fig:troudeau_prediction_ortho}
    \end{subfigure}
    \caption{Results of the UAV model applied to the entire orthophoto of the \textit{Trou d'eau} lagoon (a) and \textit{Saint-Leu} lagoon (b).
    The model predicts the presence of coral morphotypes and habitats across the entire reef area.
    Piecharts can then be computed to quantify the predicted habitat composition. 
    (c) Zoomed-in view of the UAV orthophoto of the \textit{Trou d'eau}  lagoon.
    (d) Corresponding predicted segmentation mask for the zoomed-in area of the \textit{Trou d'eau} lagoon orthophoto, showing the spatial distribution of coral morphotypes and habitats.
    Black circles highlight spurious patches of different corals predicted by the model, which are not present in the orthophoto.
    \label{fig:prediction_masks}}
\end{figure}

To assess the spatial consistency and ecological relevance of the segmentation predictions across large reef areas, we applied the trained UAV model to the entire UAV orthophotos of the \textit{Trou d'eau} and \textit{Saint-Leu} lagoons, measuring \SI{189682}{\meter\squared} and \SI{204748}{\meter\squared} respectively, see Figure \ref{fig:prediction_masks}.
As a reminder, the six ASV sessions used to train the model covered a total area of total of \SI{16904}{\meter\squared}.
Therefore, on average a single UAV session covers approximately 100 times more area than an ASV survey of equivalent time duration, assuming both platforms maintain horizontal overlap between transects.
However, this gain in spatial extent comes at the cost of resolution: the ASV imagery typically achieves a GSD of about \SI{1}{\milli\metre\per\pixel} in the lagoon, whereas UAV imagery provides a coarser resolution of roughly \SI{1}{\centi\metre\per\pixel}.

In Figures \ref{fig:troudeau_prediction} and \ref{fig:stleu_prediction}, the model predictions over the entire orthophotos of the \textit{Trou d'eau} and \textit{Saint-Leu} lagoons are displayed, accompanied by pie charts summarising the percentage cover of each predicted class.
Finally, Figure \ref{fig:troudeau_prediction_ortho} presents a zoomed-in view of the UAV orthophoto of the \textit{Trou d'eau} lagoon, while Figure \ref{fig:troudeau_ortho} shows the corresponding predicted segmentation mask.

\subsection{Garance cyclone: a case study in Réunion island}

In late February 2025, the cyclone \textit{Garance} struck Réunion Island, bringing winds of up to \SI{234}{\kilo\meter\per\hour} and extreme rainfall (\SI{140}{\milli\meter} in one hour and over \SI{250}{\milli\meter} in three hours along the north coast \citep{lamy2025climaax}).
This extreme precipitation transported large amounts of sediment from the watershed into the lagoon, depositing a thick layer of mud on coral habitats. 

\begin{figure}[ht]
    \centering
    \begin{subfigure}[t]{0.45\textwidth}
        \centering
        \includegraphics[width=\textwidth]{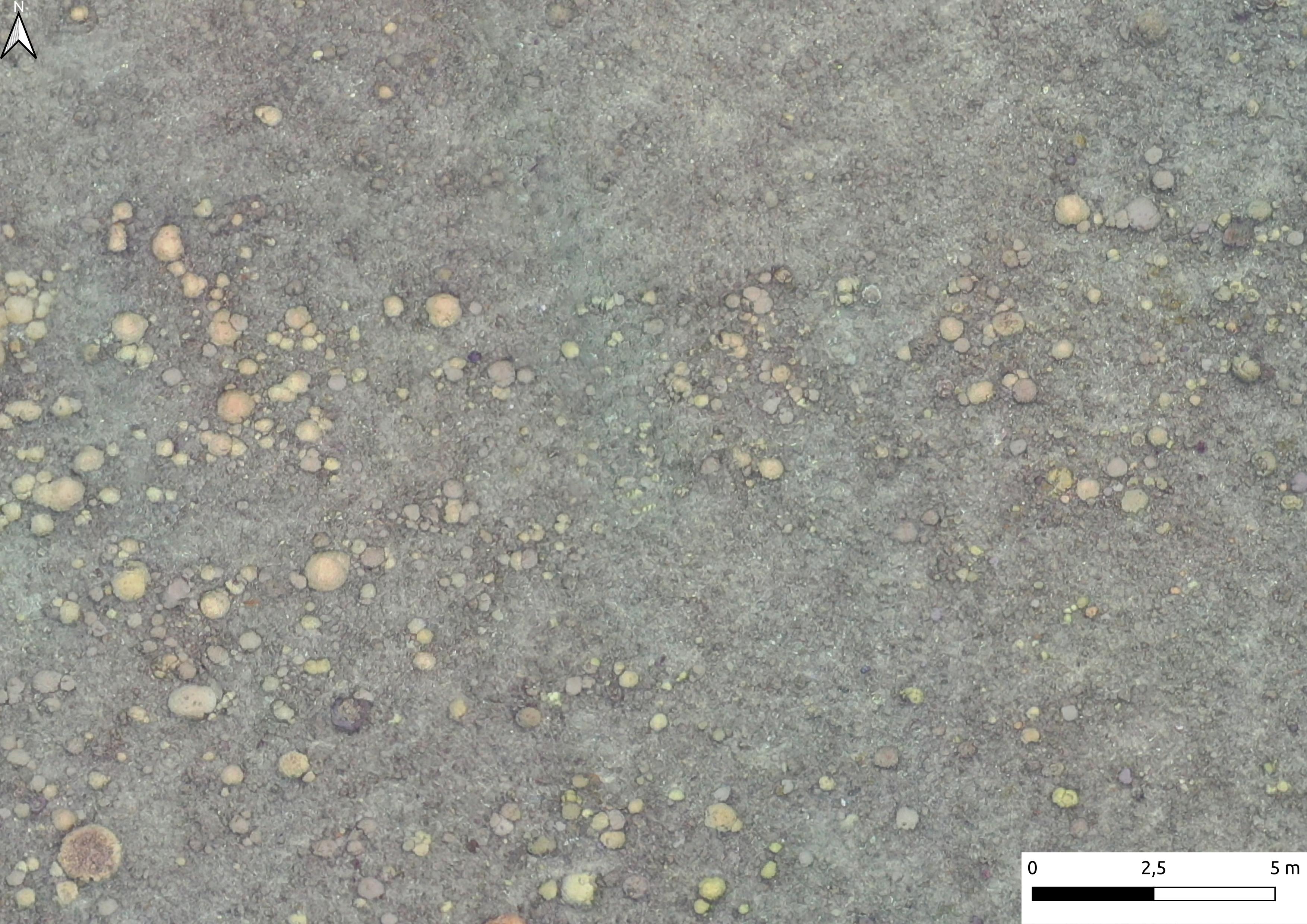}
        \caption{}
        \label{fig:ortholeu23}
    \end{subfigure}
    \hfill
    \begin{subfigure}[t]{0.45\textwidth}
        \centering
        \includegraphics[width=\textwidth]{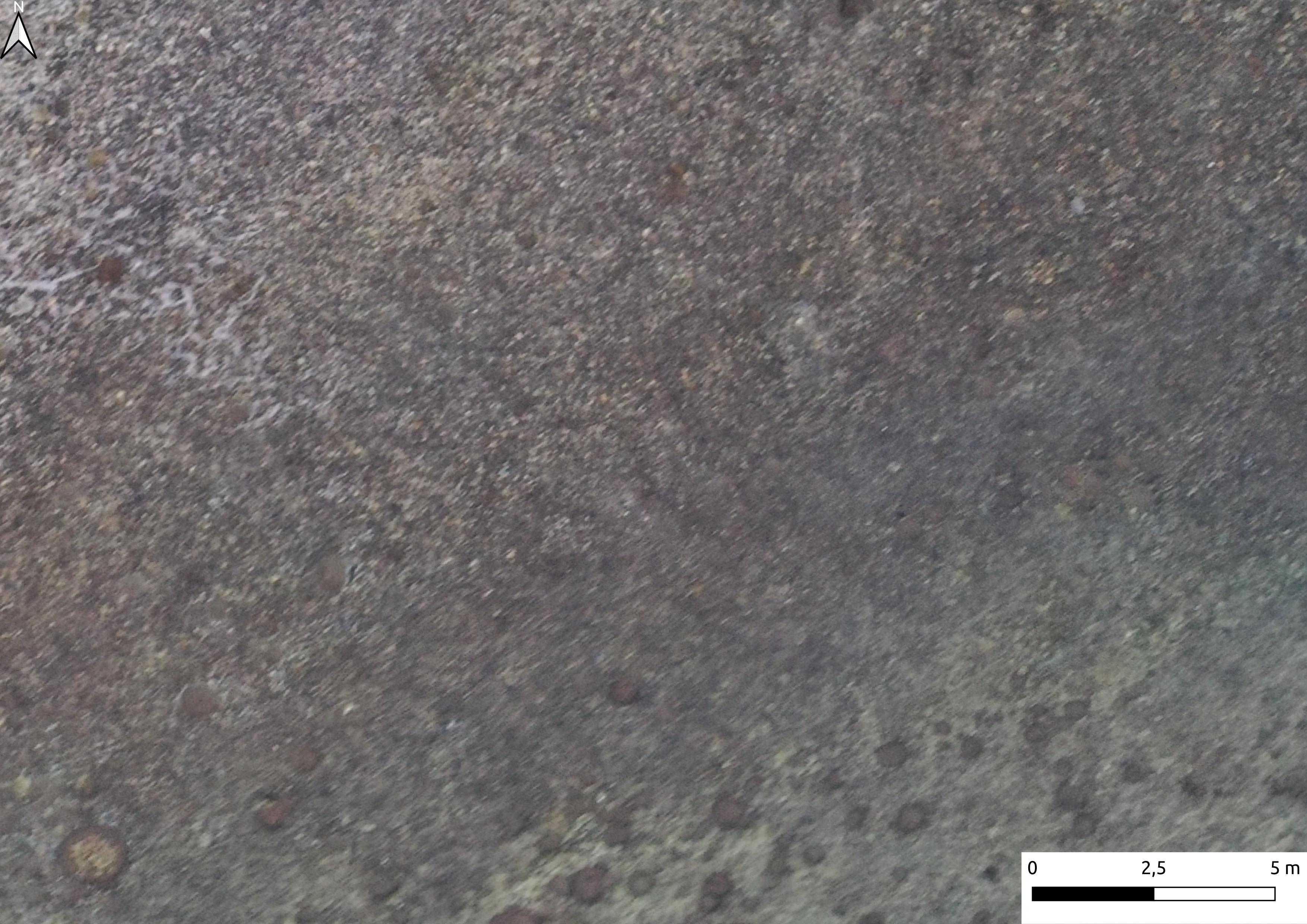}
        \caption{}
        \label{fig:ortholeu25}
    \end{subfigure}

    \vskip\baselineskip 
    
    \centering

    \begin{subfigure}[t]{0.45\textwidth}
        \centering
        \includegraphics[width=\textwidth]{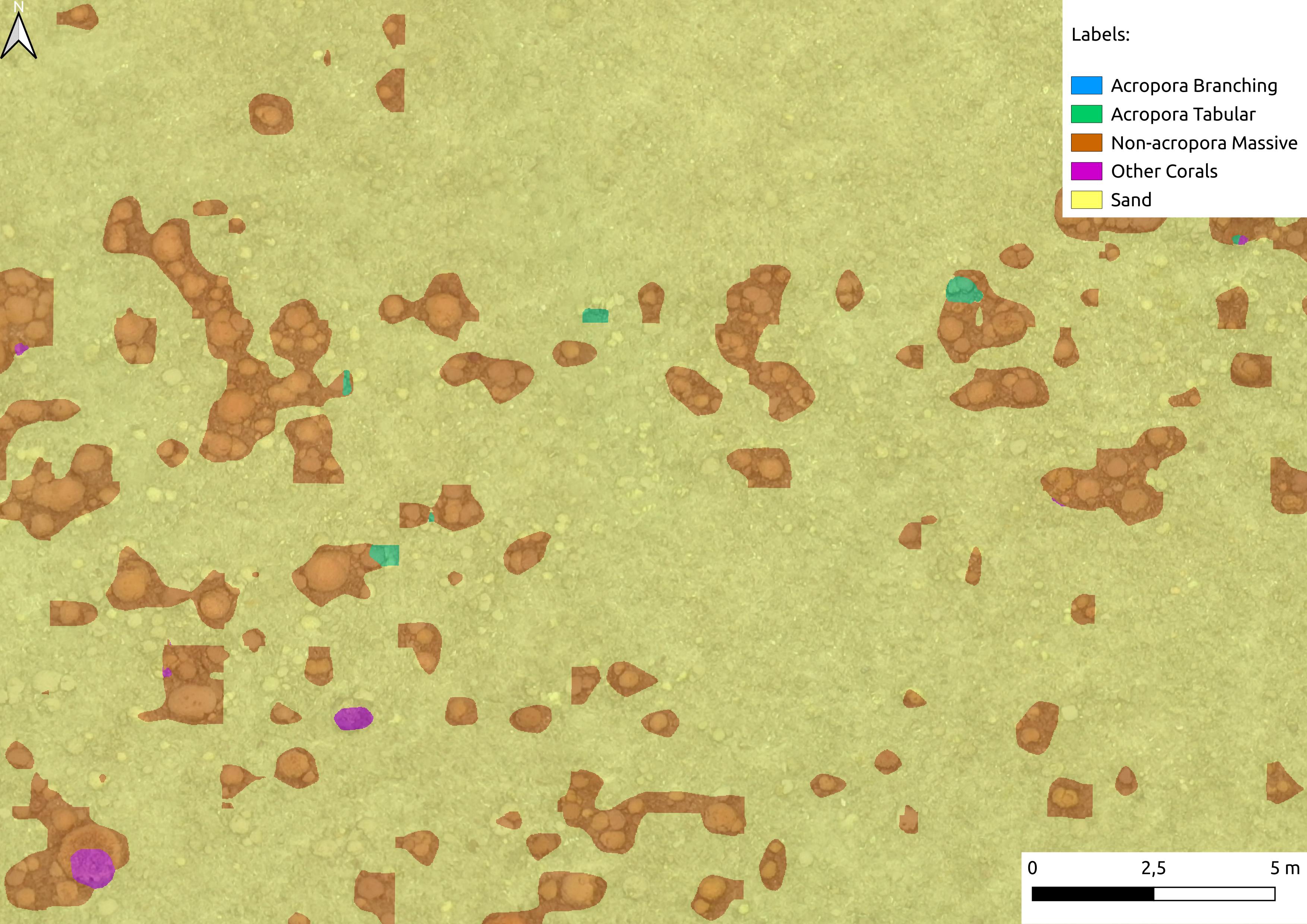}
        \caption{}
        \label{fig:predleu2023}
    \end{subfigure}
    \hfill
    \begin{subfigure}[t]{0.45\textwidth}
        \centering
        \includegraphics[width=\textwidth]{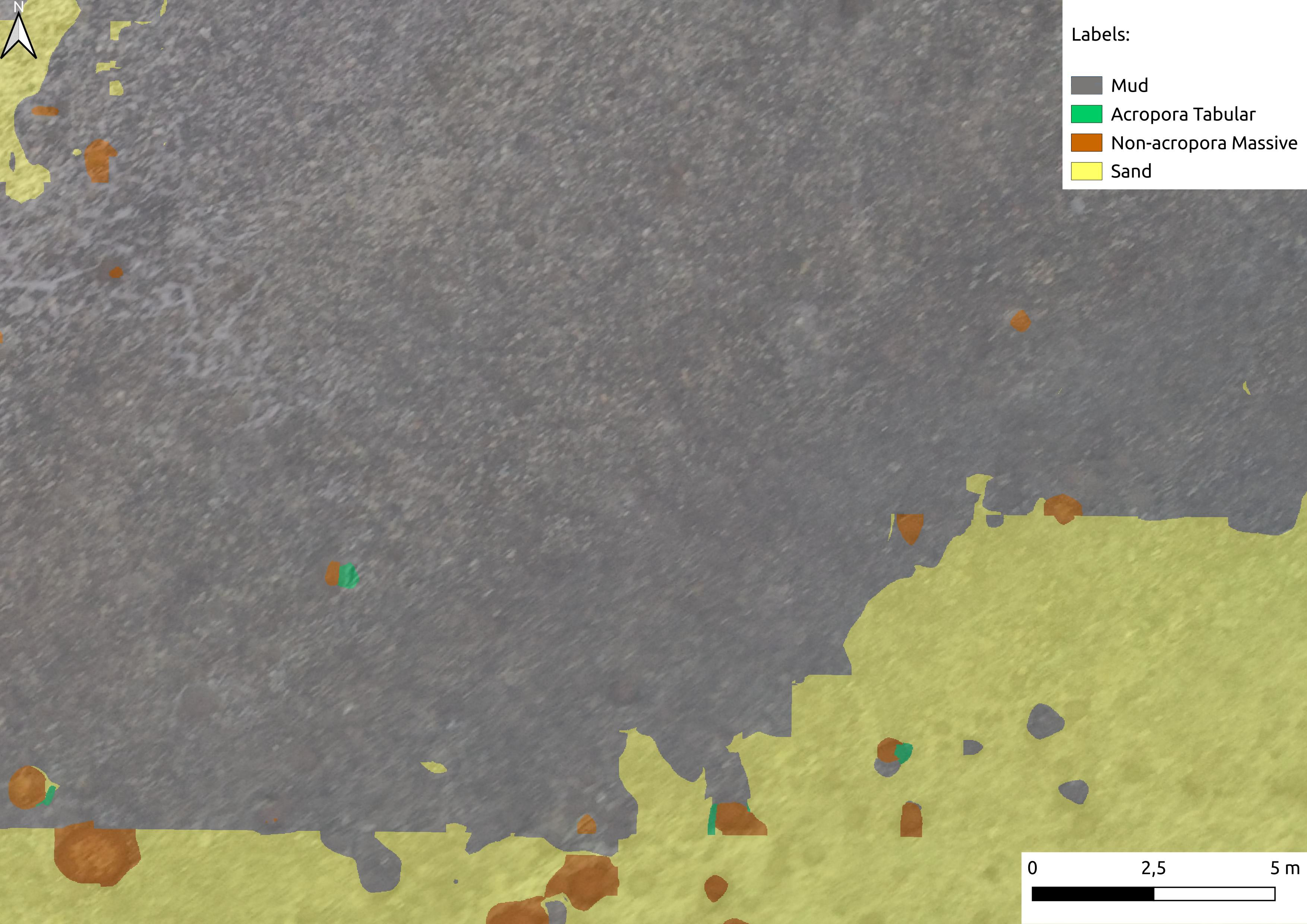}
        \caption{}
        \label{fig:predleu2025}
    \end{subfigure}
    \caption{UAV-based monitoring of the Saint-Leu lagoon before and after cyclone Garance: (a) Orthophoto from December 2023 showing a dense aggregation of massive corals. (b) Orthophoto from June 2025, four months after the cyclone, showing visible habitat changes. (c) Semantic segmentation of the 2023 orthophoto using the proposed WSSS framework, allowing identification and counting of massive corals. (d) Semantic segmentation of the 2025 orthophoto, highlighting the loss and degradation of coral cover in the same area (the grey area corresponds to mud cover).}
    \label{fig:garance_monitoring}
\end{figure}

Using the UAV-based monitoring technique described in this study, we compared two drone orthophotos acquired two years before and four months after the cyclone (Figure \ref{fig:garance_monitoring}) to identify the areas most affected.

\section{Discussion}
\label{sec_discussion_chap4}
The results of the multi-scale upscaling process demonstrate the feasibility of training a UAV-based coral segmentation model using weak annotations derived from ASV-based image classification.
As shown in Figure \ref{fig:prediction_masks}, the benthic coverage for each class can be quantified while highlighting spatial patterns throughout the reef scape.
In Figures \ref{fig:troudeau_prediction} and \ref{fig:stleu_prediction}, the model predictions over the entire orthophotos of the \textit{Trou d'eau} and \textit{Saint-Leu} lagoons are displayed, accompanied by pie charts summarising the percentage cover of each predicted class.
Although surface area is a useful metric for many benthic classes, it may be less informative for others (e.g., $\NoAcroporeM$ or mobile species) where the number of detected individuals serves as a more ecologically relevant indicator.
As demonstrated in Section \ref{subsec_moving_species}, the proposed method also enables individual-level detection and counting for such classes.
Figures \ref{fig:troudeau_ortho} and \ref{fig:troudeau_prediction_ortho} present a zoomed-in view of a specific area in the \textit{Trou d'eau} lagoon and the corresponding segmentation mask.
Black circles highlight spurious patches of different corals predicted by the model, which are not present in the orthophoto.
While minor misclassification errors can be observed, such as isolated spurious patches of $\NoAcroporeM$, the overall quality of the segmentation is encouraging.
The predicted masks accurately reflect the structure and extent of the main coral morphotypes, with clear and coherent shapes that align well with the visible habitat patterns in the aerial imagery. 
This qualitative assessment highlights the potential of the model to support spatial analyses of reef composition and guide targeted conservation actions.

\subsection{WSSS loss function}
\label{subsec_wsss_loss_function}
When dealing with WSSS the choice of the loss function can have a significant impact on the model performance.
Indeed during training, the loss function is used at each iteration to compute the error between the predicted segmentation mask and the ground truth mask.
Since the ground-truth masks are generated from weak annotations, they contain noise and inaccuracies, which must be taken into account when choosing the loss function.
For the first training of Segformer model, we tried four different loss functions: boundary loss, dice loss, cross-entropy loss and focal loss.

\begin{figure}[ht]
    \centering\includegraphics[width=0.9\textwidth]{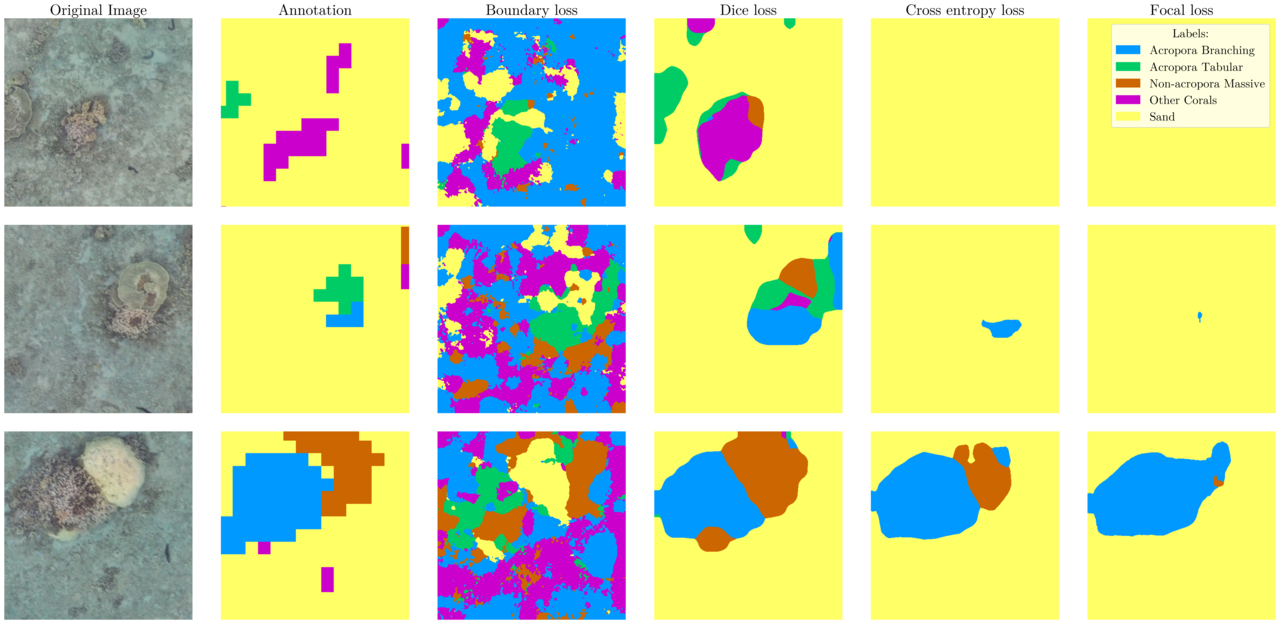}
    \caption{Comparison of four different loss functions for WSSS training with coarse annotations.
    The first column shows the original UAV tile, followed by the coarse annotation used for training.
    The remaining columns display the predicted segmentation masks obtained using boundary loss, dice loss, cross-entropy loss and focal loss, respectively.}
    \label{fig:losses_comparison}
\end{figure}

The comparison between the coarse annotation mask and the prediction masks obtained by training the model shows substantial differences between the four different loss functions (Figure \ref{fig:losses_comparison}).
The cross-entropy loss and focal loss are pixel-level losses that focus on individual pixels \citep{azad2023loss}, making them sensitive to label noise and class imbalance.
This is reflected in the predictions, that often do not recognise under-represented classes such as $\AcroporeT$ and $\NoAcroporeM$.
Boundary level losses, like the boundary loss, are designed to focus on the edges of objects, which can be useful for fine-grained segmentation tasks.
For WSSS however, boundary losses are too sensitive to noise in the annotations and this is clearly reflected in the predictions in Figure \ref{fig:losses_comparison}, which exhibits high pixelation effects.

Finally dice loss \citep{sudre2017generalised}, although sometimes produces noisier masks than the ones obtained with other losses, is the only one that correctly handled the class imbalance.
In fact, the dice loss is a region-based loss that evaluates the overlap between predicted and annotated regions as a whole, rather than focusing on individual pixels or precise edges.
This makes the loss robust to class imbalance even if less sensitive to fine edge details or precise boundary alignment.

\subsection{Aerial model evaluation}
\label{subsec_aerial_model_evaluation}
With a total accuracy of 86.07\% and a mean IoU of 52.23\%. (see Table \ref{tab:iou_by_zone}), the model effectively captures the spatial distribution of coral morphotypes and habitats in aerial imagery.
The difference between the two metrics indicates that the model is biased towards the most common class, $\Sand$, which is easily recognisable and segmentable in aerial images.
Its distinct spectral signature and broad spatial coverage in aerial imagery likely contributed to the high recall of 95.5\% (see Figure \ref{fig:confusion_matrix}).
The model also performed well on the three coral classes $\AcroporeB$, $\AcroporeT$ and $\NoAcroporeM$, correctly identifying 73.1\%, 89.3\% and 94.3\% of the annotated pixels for each class respectively.

The lowest performance is encountered for the \textit{Other Corals} class.
This class, grouping various coral morphotypes, is characterised by high intra-class variability and spectral ambiguity, leading to a lower performance of 41.3\%.
For example the $\AcroporeD$ morphotype, which is a small coral colony being part of this class, can be easily confused on aerial images with $\AcroporeB$ or $\AcroporeT$ morphotypes, which explains the high misclassification rates of 36\% and 6.5\% respectively, as shown in Figure \ref{fig:confusion_matrix}.
Furthermore, the \textit{Other Corals} class groups a wide variety of coral morphotypes that differ greatly in appearance and structure.
For instance, it includes both large, dome-shaped colonies that resemble massive corals and also flat, spreading forms like encrusting corals, which are morphologically and spectrally distinct.
This high intra-class variability makes it difficult for the model to learn coherent and consistent features, reducing its classification performance.

To further assess the robustness of these results, we estimated 95\% confidence intervals using a spatial block bootstrap over the annotated test zones (see Table \ref{tab:bootstrap_ci}).
This procedure accounts for spatial correlation between neighbouring pixels and provides a more realistic estimate of performance variability than pixel-wise assumptions of independence.
The resulting confidence intervals confirm the overall robustness of the model while highlighting variability across classes.
In particular, dominant classes such as $\Sand$ exhibit narrow confidence intervals, reflecting stable predictions across spatial samples.
In contrast, less represented classes such as $\AcroporeT$ and $\NoAcroporeM$ show wider intervals, indicating higher sensitivity to local spatial variability and annotation sparsity.
These results suggest that while the model provides reliable large-scale mapping of dominant benthic habitats, performance on minority or visually complex classes remains more uncertain.
This limitation is consistent with both class imbalance and the intrinsic variability of coral morphotypes.

While the current model was not trained to recognize the full spectrum of coral morphotypes, the methodology itself is generic and has been validated for its ability to incorporate new classes as needed (see Section \ref{subsec_moving_species}).
Although subdividing the \textit{Other Corals} category into more specific morphotypes would likely improve accuracy, this was not pursued in the present study due to the need for more detailed annotations, an expanded underwater dataset and higher-resolution UAV imagery to ensure reliable training.

\subsection{Expanding to new ecological classes}
\label{subsec_moving_species}

The presented methodology can be easily extended to include additional benthic classes, such as specific types of seagrass or other coral morphotypes not found in Réunion Island.  
To incorporate new classes, it is sufficient to collect both underwater and aerial images in a region where the new class is present and to train the student model using the ASV-derived annotations following the same upscaling methodology.
It is important to note that the method remains applicable regardless of the specific architecture of the underwater model, as long as it provides reliable class-level predictions (whether from monolabel, multilabel or segmentation classifiers).
This flexibility ensures the continued relevance and adaptability of the approach as artificial intelligence models evolve, while also enabling its transferability to different reef systems for broader regional monitoring without requiring changes to the core pipeline.

Another way to incorporate new species, including mobile ones such as sea cucumbers or turtles, is to manually annotate a small number of instances directly in the aerial imagery.
For slow-moving species such as sea cucumbers, the collection of fine-scale ASV images and UAV images simultaneously can provide the spatial and temporal alignment required to propagate class labels from underwater predictions to aerial images \citep{contini2025underwater}.
However, weather conditions or logistical constraints may prevent synchronised data collection.  
In addition, for fast-moving species such as turtles, simultaneous acquisition is not feasible.
In these situations, a few manual annotations can be added on top of ASV-generated masks to introduce new classes to the training dataset.

The introduction of SAM in 2023 by \citet{kirillov2023segment} has greatly simplified the annotation process.  
Tools such as the \textit{Geo-SAM} plugin for QGIS \citep{zhao2023geo} allow users to create object masks by clicking on points or drawing bounding boxes, reducing the time required for detailed labelling.  
For instance, using Geo-SAM, we manually annotated 338 sea cucumbers in a \SI{13840}{\meter\squared} area of the \textit{Trou d'eau} lagoon in approximately 55 minutes.

These annotations were integrated with the existing model predictions in the training zone described in Section \ref{sec_underwater} by overlaying them onto the predicted masks.
The student model was then retrained from scratch to include the new $\SeaC$ class.
Given that sea cucumbers are small organisms occupying a limited number of pixels in aerial images, their representation in the training data was proportionally low.
To address this imbalance and ensure the model could effectively learn to detect them, we adjusted the loss function by applying a weighted Dice loss, assigning a higher penalty (seven times the weight of other classes) to errors involving sea cucumbers.
The training process did not involve fine-tuning from previous weights to avoid bias toward earlier class distributions and to assess whether the inclusion of new categories interferes with performance on existing ones.

The Segformer \textit{B0} model was tested on the same test zones as before, specifically the \textit{Trou d'eau} and \textit{Saint-Leu} lagoons, by adding the $\SeaC$ class to the ground truth annotation masks.
Model evaluation showed that the new class was detected with high accuracy:
\begin{itemize}
  \item Pixel Accuracy: 0.7545
  \item IoU : 0.4817
\end{itemize}
demonstrating the model's ability to learn from a small number of manual annotations.

To assess whether introducing $\SeaC$ impacted the model's performance on other classes, we compared standard metrics with those from the original 5-class version.
As shown in Table \ref{tab:iou_seacucumber}, a decrease in overall accuracy (from 85.72\% to 83.35\%) and mean IoU (from 50.10\% to 44.48\%) was observed, indicating a modest reduction in segmentation performance after the addition of the new class.

\begin{table}[ht]
\small
\centering
\caption{IoU per class, total accuracy and mean IoU with the extended 6-class model including $\SeaC$.}
\label{tab:iou_seacucumber}
\resizebox{\columnwidth}{!}{%
\begin{tabular}{>{\scriptsize}l >{\scriptsize}c >{\scriptsize}c | >{\scriptsize}c >{\scriptsize}c >{\scriptsize}c >{\scriptsize}c >{\scriptsize}c >{\scriptsize}c}
    \toprule
    Zone & Accuracy & Mean IoU &
    Acropora B. & Acropora T. & Non-acro M. & Other Corals & Sand & Sea Cucumber \\
    \midrule
    \textbf{Total} & \textbf{0.8335} & \textbf{0.4448} & 
    \textbf{0.4638} & \textbf{0.1560} & \textbf{0.3905} & \textbf{0.2511} & \textbf{0.9254} & \textbf{0.4817} \\
    \bottomrule
\end{tabular}
}
\end{table}

The IoU scores for all coral-related classes declined: for \textit{Other Corals} (from 0.4111 to 0.2511) and $\AcroporeT$ (from 0.2313 to 0.1560). 
Similarly, $\AcroporeB$ and $\NoAcroporeM$ saw moderate reductions (from 0.5282 to 0.4638 and 0.4095 to 0.3905, respectively).
These drops suggest that the integration of a new, visually distinct class, such as $\SeaC$, may have introduced some degree of class interference, likely due to the high penalty weighting assigned to it in the loss function.
Importantly, the performance on the $\Sand$ class remained stable (0.9249 vs 0.9254), suggesting that well-represented, visually dominant classes are more robust to such extensions.

\begin{figure}[ht]
    \centering
    \begin{subfigure}[t]{0.45\linewidth}
        \centering
        \includegraphics[width=\linewidth]{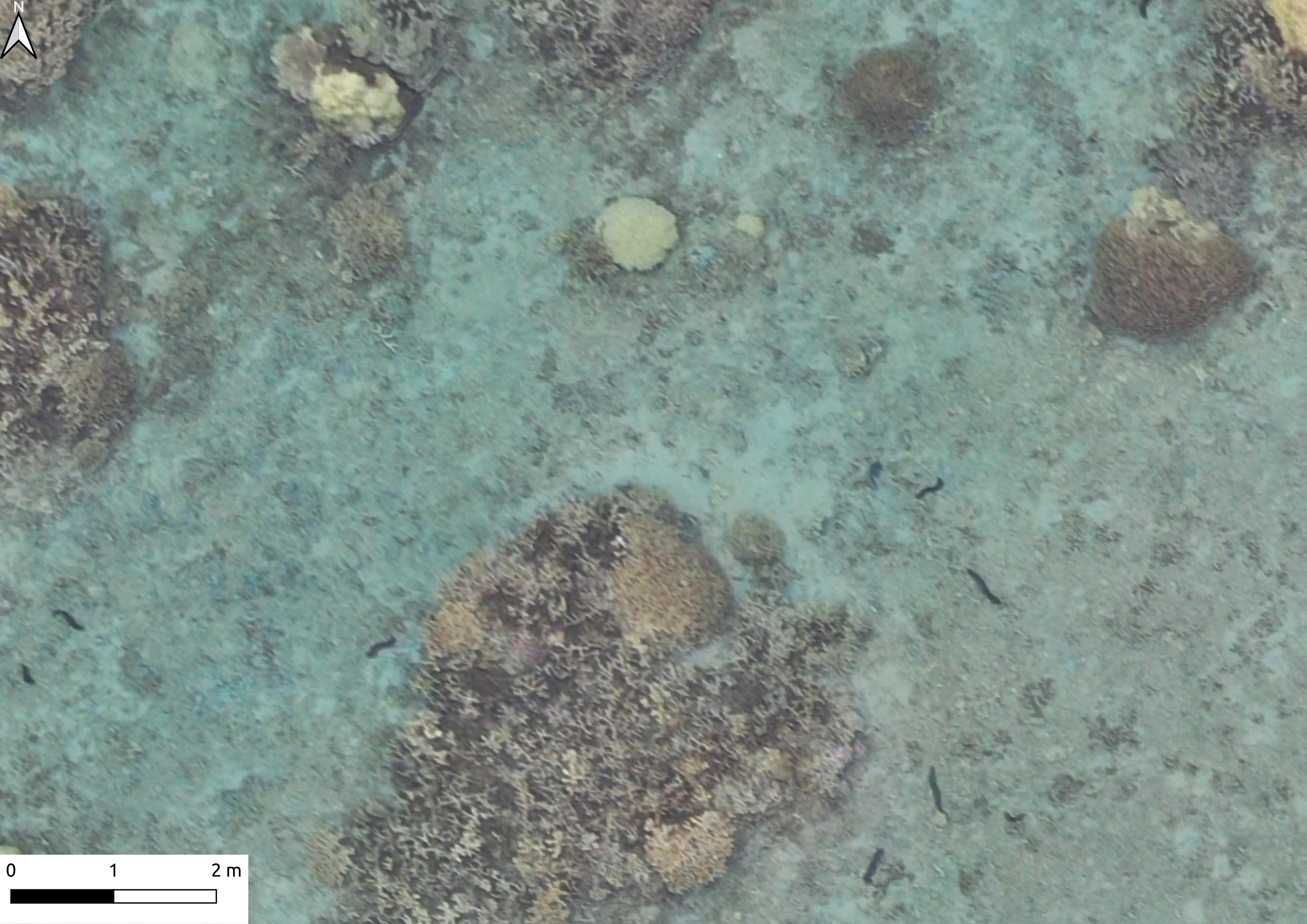}
        \caption{}
        \label{fig:sea_cucumber_ortho}
    \end{subfigure}
    \hfill
    \begin{subfigure}[t]{0.45\linewidth}
        \centering
        \includegraphics[width=\linewidth]{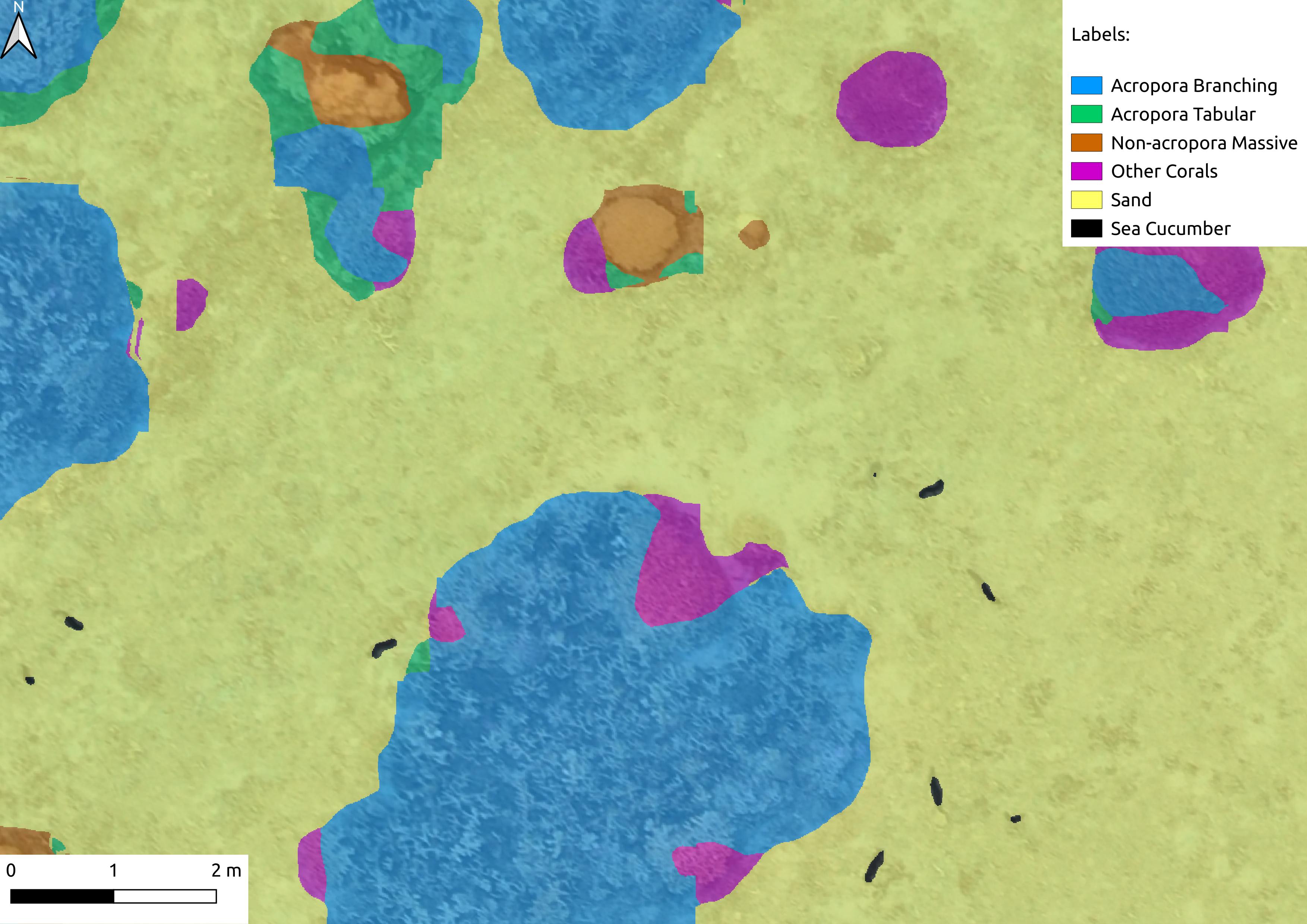}
        \caption{}
        \label{fig:sea_cucumber_prediction}
    \end{subfigure}
    \caption{Example of the segmentation output for the $\SeaC$ class.
    (a) Zoomed-in view of the UAV orthophoto of the \textit{Trou d'eau} lagoon showing the presence of sea cucumbers.
    (b) Corresponding predicted segmentation mask, showing the spatial distribution of sea cucumbers.}
    \label{fig:sea_cucumber_predictions}
\end{figure}

Figure \ref{fig:sea_cucumber_ortho} illustrates a zoomed-in view of the UAV orthophoto of the \textit{Trou d'eau} lagoon showing the presence of sea cucumbers and in Figure \ref{fig:sea_cucumber_prediction} the corresponding predicted segmentation mask.
Some key ecological information can be extracted from such maps, including the spatial density of individuals per square metre and estimates of the size of the organism.
For instance, the average length of each detected sea cucumber can be approximated by measuring the longest axis within each predicted segmentation mask.
In the back reef depression of the \textit{Trou d'eau} lagoon, this method yielded an average length of $19.6 \pm 0.24$ cm, calculated over 742 predicted individuals.
These estimates are consistent with \textit{in situ} measurements reported in \citet{pierrat2024searching}, where average lengths of $24.6 \pm 0.5$\,cm and $12.5 \pm 0.7$\,cm were recorded for the sea cucumber species \textit{Holothuria leucospilota} and \textit{Stichopus chloronotus}, respectively.  
Based on 72 individuals per species (144 in total), this yields a combined average length of $18.5$ cm, in close agreement with our estimate.

\subsection{Long term reef monitoring}
\label{subsec_longterm}

The ecological dimension of the framework is currently constrained by the absence of long-term standardized monitoring at the same sites. 
As a result, it is not yet possible to systematically quantify reef ecosystem dynamics or to rapidly detect ecological changes caused by disturbances such as cyclones, bleaching events or human impacts.

A key perspective is the long-term implementation of standardized monitoring protocols, enabling repeated surveys of the same areas.
This temporal approach would facilitate the rapid detection of changes, the quantification of ecosystem trends and the provision of robust data for conservation management. 

Recent advances in UAV technology have produced fully autonomous multi-rotor platforms capable of inspecting large sites without human intervention \citep{knitter2024survey}.  
When paired with hangar systems, these drones can autonomously land, recharge and relaunch, enabling continuous operation depending on local regulatory constraints.

An example of such a monitoring technique is shown in Figure \ref{fig:garance_monitoring}, where the WSSS workflow described in this study, is applied to a zone directly in front of two ravines converging under a bridge where UAV orthophotos collected two years before and four months after the cyclone (Figure \ref{fig:garance_monitoring}) were available.
The 2023 orthophoto prior to cyclone \textit{Garance}, shown in Figure \ref{fig:ortholeu23} and covering an area of \SI{534.218}{\square\metre}, depicts an area occupied by numerous massive corals part that can be distinguished and individually identified (Figure \ref{fig:predleu2023}).
In the post-cyclone orthophoto (Figure \ref{fig:ortholeu25}), a thick mud layer covers much of the surveyed area, burying a large number of massive corals.
In the lower portion of the image, however, several massive corals remain visible.
In the corresponding segmentation (Figure \ref{fig:predleu2025}), these remaining colonies are correctly recognised, while areas covered by mud are frequently misclassified. 
For illustration purposes, this misclassification has been coloured in grey as it reflects the absence of sediment-covered examples in the training dataset.
This drawback could be addressed by adding a dedicated \textit{Mud} class, as described in Section \ref{subsec_moving_species}.
Despite this limitation, the framework effectively identifies coral loss: 89.70\% of $\NoAcroporeM$ colonies present in 2023 were no longer detected in the 2025 survey.
In marine conservation, these systems could be deployed for periodic lagoon surveys, providing high-frequency observations that enable the early detection of ecological changes.

\subsection{Comparing multi-label classification and semantic segmentation approaches}
\label{subsec_comparison}

In \citep{contini2025underwater}, a multi-label classification framework was adopted to transfer knowledge from ASV-based underwater predictions to UAV-based aerial imagery.
This method associates each aerial tile with a list of predicted classes, indicating which coral morphotypes or habitats are likely present in the image, without specifying their exact location.
In contrast, the current study introduces a semantic segmentation approach, where the model must assign a class label to every pixel in the image.

One of the main advantages of the multi-label classification framework is its flexibility in data collection.
Since this method requires only the presence or absence of classes at the tile level, the image dataset does not need to be densely distributed over a specific area.
As long as the underwater images are collected across representative zones and sufficiently cover each aerial tile, the model can learn to infer class presence on UAV orthophotos.
This makes the upscaling process easier to deploy, especially when fine scale data is limited or hard to collect.
In comparison, the semantic segmentation approach demands a much denser distribution of underwater images over the area of interest.
This is because generating weak segmentation masks from ASV predictions involves interpolating spatially continuous rasters.
To reliably locate where each class appears within a tile, a complete and high-resolution coverage of the zone is recommended.

Another important difference lies in how class context is transferred from underwater to aerial imagery.
In multi-label classification, the aerial model can learn to associate contextual information with the presence of a class, even if the class is not clearly visible on the image.
For example, certain coral types like $\NoAcroporeE$ may not be directly visible in the aerial image, but their presence can be inferred from surrounding textures, structures, or typical spatial patterns at the tile level.
Semantic segmentation, on the other hand, requires the model to assign a class to each pixel.
This makes it harder to transfer class presence if the class is not clearly visible, since there is no straightforward way to represent contextual cues in pixel-wise annotations.

Despite these limitations, semantic segmentation provides a clear advantage in terms of spatial details since it allows for precise mapping of coral morphotypes and enables the computation of spatial coverages or counting of marine organisms.
This level of detail is essential for certain ecological applications, such as tracking coral recovery \citep{zhang2022deep}.
Furthermore, as demonstrated in Section \ref{subsec_moving_species}, segmentation masks enable the extraction of additional metrics beyond simple presence/absence, such as organism density and size distributions.

In summary, the multi-label classification approach offers greater flexibility and simplicity in deployment, especially in scenarios with sparse underwater data.
Semantic segmentation, while more demanding in terms of data and annotation requirements, enables richer and more spatially explicit predictions that are better suited for detailed ecological analyses.
These two approaches are therefore complementary and the choice between them depends on the monitoring objectives and data availability.

\subsection{Limits and future directions} 
\label{sec_future}
The proposed pipeline is inherently dependent on the quality of the teacher model used to generate weak annotations. 
As the \textit{DinoVDeau} model provides the initial supervision through ASV-based predictions, its errors (e.g., false positives, missed detections, or class confusion) are directly propagated into the interpolated rasters and then into the training data of the UAV segmentation model. 
Future work could therefore explore the use of more recent foundation models for underwater image classification, which may provide more accurate and robust predictions and thereby improve the quality of the generated supervision. 
In particular, recent advances in computer vision, such as DINOV3 \citep{simeoni2025dinov3}, suggest that significant performance gains can be achieved by leveraging stronger feature representations. 
Such developments indicate that the proposed framework could naturally benefit from ongoing progress in representation learning, without requiring fundamental modifications to its design.

A second limitation arises from the interpolation and upsampling steps used to generate training annotations. 
Interpolation induces spatial smoothing, which may blur fine-scale structures and spread local classification errors over neighbouring areas, particularly in regions with sparse sampling. 
However, in our setting, the upsampling step plays a more critical role due to the strong resolution mismatch between ASV-derived rasters (on the order of tens of centimetres) and UAV orthophotos (centimetre-level resolution). 
The required upsampling of coarse rasters to high-resolution grids may lead to block artefacts and loss of boundary precision, limiting the spatial accuracy of the generated annotations. 
Future work could address this limitation by increasing the spatial density of underwater image acquisition, thereby reducing the scale gap between ASV-derived data and UAV imagery. 

\begin{figure}[ht]
         \centering  \includegraphics[width=0.6\linewidth]{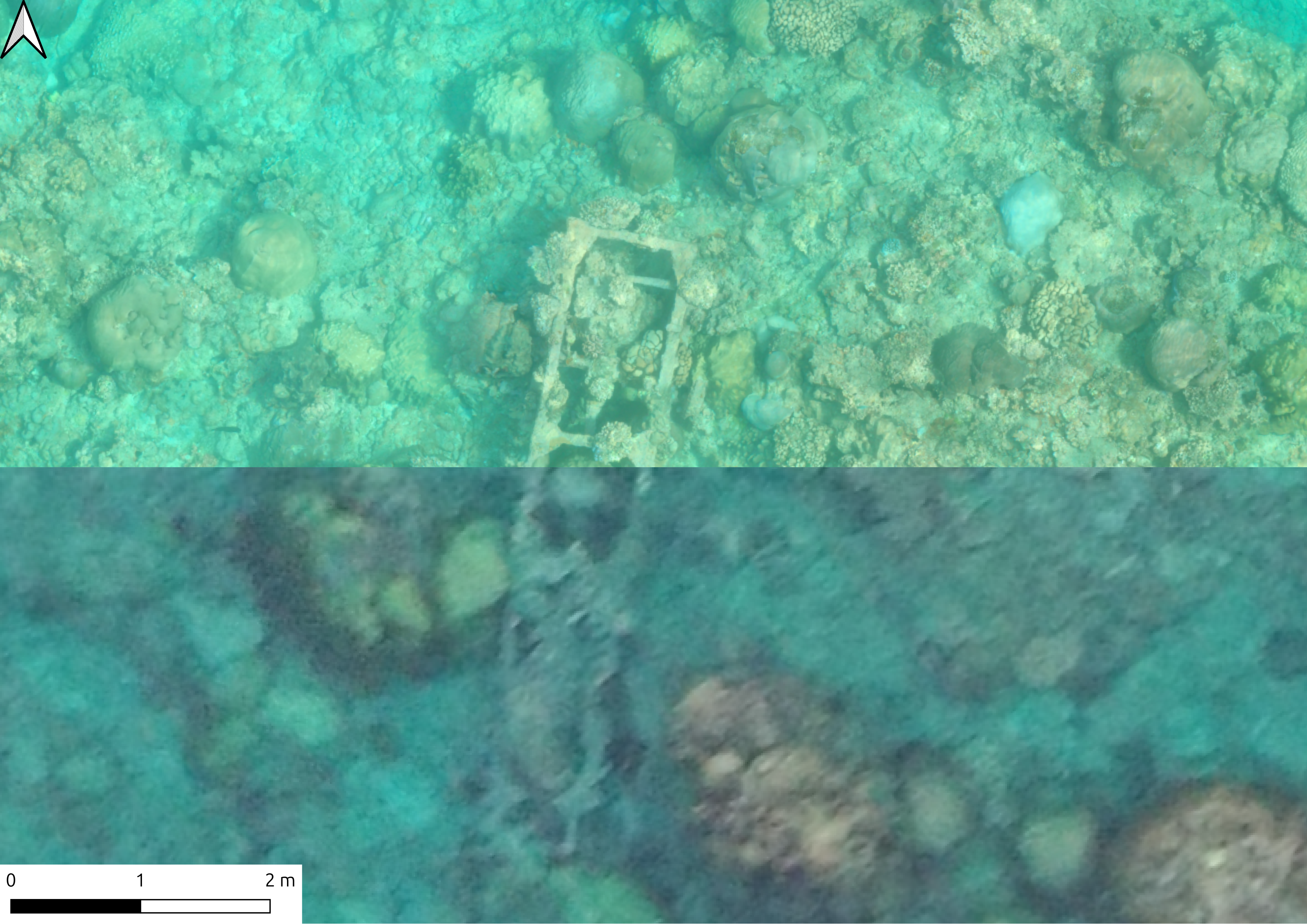}
         \caption{Comparison between underwater (top) and aerial (bottom) orthophotos of the same reef zone at Saint-Leu (Réunion Island) illustrating the difference in level of detail.}
    \label{fig:ortho_underwater_and_aerial}
 \end{figure}

From a technical standpoint, the framework remains limited by the difficulties of producing reliable underwater orthophotos, the high annotation costs 
associated with semantic segmentation and the current dependence on multi-label classification rather than fully pixel-wise approaches. 
These constraints reduce the spatial precision of the annotations transferred from underwater predictions to the aerial scale and restrict the number of benthic classes that can be included in the analysis.
Recent advances in underwater photogrammetry and zero-shot segmentation models (such as SAM) now make the construction and annotation of these orthophotos considerably more feasible and, as public underwater segmentation datasets continue to grow \citep{truong2025msc, sauder2025coralscapes}, the availability of high-resolution underwater segmentation models increases \citep{remmers2025rapidbenthos}.
This evolution opens the possibility of transferring high-resolution semantic masks from underwater to aerial models, enabling two key improvements:

\begin{enumerate}
    \item Spatially precise information transfer: 
    as illustrated in Figure \ref{fig:ortho_underwater_and_aerial}, correctly aligned underwater orthophotos can provide detailed information that can be transferred to aerial imagery.  
    The upper part of the figure shows an underwater orthophoto reconstructed from ASV-acquired frames, revealing fine-scale habitat details.  
    The lower part displays the corresponding aerial orthophoto obtained with an UAV, which offers broader coverage but at lower resolution.  
    By transferring underwater prediction masks to the aerial scale (rather than relying on centroid-based annotations for each frame) it becomes possible to generate full segmentation masks that capture the exact boundary of each instance in the underwater image, thereby preserving the spatial distribution of benthic classes.
        
    \item Increased class diversity:  
    pixel-wise underwater annotations allow overlapping or co-occurring classes to be distinguished within the same image.  
    For example, instead of assigning a single dominant label such as $\Sand$ to an entire frame that also contains patches of $\Rubble$, segmentation masks can isolate the specific pixels corresponding to each substrate.  
    This refinement enables the transfer of richer and more precise ecological information to the aerial scale, making it possible to include classes such as \textit{Algae}, $\Rubble$ or coral morphotypes that were previously excluded due to annotation noise.

\end{enumerate}

As the technological landscape continues to evolve, several parallel advancements further motivate and reinforce the relevance of this line of research.
Advancements in UAV hardware are among the most promising.
Modern drones now feature higher-resolution sensors, enhanced optics and differential Global Navigation Satellite System, enabling more precise georeferencing and faster acquisition of imagery over larger areas.
The combination of better spatial resolution and accurate positioning improves the ability to detect and distinguish ecologically meaningful classes in aerial imagery, facilitating finer-scale and more reliable large-area habitat mapping.

Finally, state of the art models are now tackling increasingly complex tasks, including 3D point cloud segmentation and cross-domain multimodal fusion, which could allow direct analysis of combined aerial and underwater 3D datasets.
These developments indicate that multi-scale, multi-platform frameworks like the one proposed in this thesis are likely to benefit substantially from upcoming AI-driven methods, ultimately enabling frequent, high-resolution and cost-effective environmental monitoring at unprecedented spatial scales.

\section{Conclusion} 
\label{sec_conclusion_chap4}
This study introduces a multi-scale WSSS framework for coral reef monitoring that transfers fine-scale information from underwater imagery to large-scale aerial assessments. 
Using dense ASV image collections and a deep learning classification model, we generated spatially interpolated probability maps that serve as weak annotations for training UAV-based segmentation models, eliminating the need for pixel-level labels.

Despite coarse supervision, the method enables accurate segmentation of key benthic classes and captures ecologically relevant spatial patterns.
Region-based loss functions and model refinement via self-distillation further improve segmentation performance.

The approach is flexible and extensible.
New classes can be incorporated either by collecting a small number of additional ASV sessions in areas where these classes are present or by manually annotating a limited number of instances directly in aerial imagery, as demonstrated for sea cucumbers.

Beyond UAV-based coral mapping, this multi-scale knowledge transfer strategy offers broader potential for hierarchical remote sensing.
The same approach could be extended to transfer information from UAV orthophotos to airborne or satellite imagery, enabling the creation of large-scale coral reef maps that combine local ecological detail with regional coverage.

This framework offers a practical and cost-effective solution for ecological monitoring in complex and changing marine environments. 
By reducing annotation requirements and leveraging multi-scale data sources, it contributes to the development of robust, scalable tools for automated coral reef assessment.

\section{Data availability} 
\label{sec_data}
The data that support the findings of this study have been assigned DOIs and are openly available (see  
\href{https://zenodo.org/records/11125848}{Seatizen Atlas} record \citep{matteo_contini_2024_13374497} on Zenodo) 
and described in a dedicated data article \cite{contini2025seatizen}.

All code required to reproduce the experiments is publicly available at 
\href{https://github.com/SeatizenDOI/the-point-is-the-mask}{the-point-is-the-mask Github} and has also been assigned a DOI on \href{https://doi.org/10.5281/zenodo.15455744}{Zenodo} and \href{https://archive.softwareheritage.org/browse/origin/?origin_url=https://doi.org/10.5281/zenodo.15455743}{Software Heritage}. 
The repository includes scripts for data downloading, preprocessing, raster generation, model training and evaluation, along with all required configuration files and dependencies.
The full pipeline can be executed directly using the provided training scripts, enabling reproducible results across different systems without additional setup.
The ASV and UAV sessions used in this study can be automatically retrieved using the provided tools, either via the 
\href{https://github.com/SeatizenDOI/zenodo-tools}{zenodo-tools Github} or directly through the main repository.

\section{Code availability} 
\label{sec_code}
All code for data processing associated with the current submission is available on \href{https://github.com/SeatizenDOI/the-point-is-the-mask}{the-point-is-the-mask Github}.

The code for downloading data associated with the current submission is available on \href{https://github.com/SeatizenDOI/zenodo-tools}{zenodo-tools Github}.
The code used to train the multi-label image classification neural network model used in the current submission \cite{DinoVdeau_teacher} is available on \href{https://github.com/SeatizenDOI/DinoVdeau}{DinoVdeau Github}.

\section{Declaration of Generative AI and AI-assisted technologies in the writing process}
During the preparation of this work the author(s) used ChatGPT-4o and GitHub Copilot in order to improve language and readability.
After using this tool/service, the author(s) reviewed and edited the content as needed and take(s) full responsibility for the content of the published article.
This tools were not involved in the design, implementation, data analysis, or manuscript preparation of the study.

\section{Acknowledgement}
\label{sec_acknowledgement}
The authors acknowledge the Pôle de Calcul et de Données Marines (PCDM) for providing DATARMOR storage, support services and computational resources. 
We extend our heartfelt thanks to Laurence Maurel, Cam Ly Rintz, Leanne Carpentier, Magali Duval, Laura Babet, Belen De Ana, Anne-Elise Nieblas, Arthur Lazennec, Victor Russias, Mervyn Ravitchandirane, Mohan Julien, Pierre Gogendeau, Thomas Chevrier and Justine Talpaert Daudon, involved in the annotation and collection of data for this study. 
Their contributions were indispensable to our research efforts. 
We also thank the reviewers for their insightful comments and suggestions, which helped improve the quality of this manuscript.

\section{Funding} 
\label{sec_funding}
This work was supported by several projects: Seatizen (Ifremer internal grant), Plancha (supported by the Contrat de convergence et de transformation 2019-2022, mesure 3.3.1.1 de la Préfecture de la Réunion, France), IOT project (funded by FEDER INTERREG V and Prefet de La Réunion: grant \#20181306-0018039 and the Contrat de Convergence et de Transformation de la Préfecture de La Réunion), Ocean and Climate Priority Research Programme, FISH-PREDICT project (funded by the IA-Biodiv ANR project: ANR-21-AAFI-0001-01) and G2OI FEDER INTERREG V (grant \#20201454-0018095).

\section{References} 
\label{sec_references}
\bibliographystyle{unsrtnat}
\bibliography{the_point_is_the_mask} 

\end{document}